\documentclass[]{CUP-JNL-EDS-PJ}

\usepackage{latexsym}
\usepackage{graphicx}
\usepackage{multicol,multirow}
\usepackage{amsmath,amssymb,amsfonts}
\usepackage{mathrsfs}
\usepackage{amsthm}
\usepackage{rotating}
\usepackage{appendix}
\usepackage[authoryear]{natbib}
\usepackage{ifpdf}
\usepackage[T1]{fontenc}
\usepackage{times}
\usepackage{newtxmath}
\usepackage{textcomp}%
\usepackage{hyperref}
\usepackage{lipsum}
\usepackage{lineno}
\usepackage[normalem]{ulem}
\newcommand{\pj}[1]{\textcolor{blue}{#1}}

\begin{document}

\begin{center}
\Large {\bf Uncertainty Quantification for Deep Learning}\\
\end{center}
\normalsize

\bigskip
\noindent Peter Jan van Leeuwen, J. Christine Chiu, and C. Kevin Yang\\
Department of Atmospheric Science, Colorado State University, Fort Collins, USA\\

\bigskip
Accepted by Environmental Data Science




\authormark{Van Leeuwen et al.}

\keywords{Uncertainty Quantification, Machine Learning, Deep Learning}

\abstract{We present a critical survey on the consistency of uncertainty quantification used in deep learning and highlight partial uncertainty coverage and many inconsistencies. We then provide a comprehensive and statistically consistent framework for uncertainty quantification in deep learning that accounts for all major sources of uncertainty: input data, training and testing data, neural network weights, and machine-learning model imperfections, \pj{targeting regression problems}. We systematically quantify each source by applying Bayes’ theorem and conditional probability densities and introduce a fast, practical implementation method. We demonstrate its effectiveness on a simple regression problem and a real-world application: predicting cloud autoconversion rates using a neural network trained on aircraft measurements from the Azores and guided by a two-moment bin model of the stochastic collection equation. In this application, uncertainty from the training and testing data dominates, followed by input data, neural network model, and weight variability.  Finally, we highlight the practical advantages of this methodology, showing that explicitly modeling training data uncertainty improves robustness to new inputs that fall outside the training data, and enhances model reliability in real-world scenarios.}

\nolinenumbers

\policy{This survey paper presents a critical overview of uncertainty quantification in deep learning. Key limitations are identified in capturing the full range of uncertainties encountered in real-world applications. A complete and consistent uncertainty quantification methodology is proposed. Through both simple illustrative examples and a complex real-world case study, the paper demonstrates where current methods fall short and how the proposed complete framework offers a more robust and statistically consistent alternative. Comparative results underscore the advantages of the new methodology in effectively addressing uncertainty across multiple sources. }


\section[]{Introduction and initial survey}
Many processes in the geosciences are highly complex, computationally expensive, or not fully understood. In such cases, machine learning -particularly deep learning- has emerged as a valuable tool to either replace costly numerical models or to learn relationships between variables when governing equations are unknown. For instance, radiative transfer processes are well understood but computationally expensive, whereas cloud microphysics is only partially understood and cannot be explicitly resolved at the scales used in weather or climate models. Despite promising applications, the adoption of deep learning in scientific domains has been slow, primarily due to the lack of robust uncertainty quantification, which is essential for model comparison, forecasting, and risk assessment.

Uncertainty can be defined in many ways. A common distinction in statistics is between aleatoric and epistemic uncertainty. Aleatoric uncertainty, also known as stochastic or random uncertainty, refers to inherent variability that arises each time an experiment is repeated, such as measurement noise from a well-calibrated instrument. Epistemic uncertainty, also called systematic or structural uncertainty, stems from limitations in our models or understanding, such as uncertainty related to the design of a neural network. In practice, the boundary between these two types of uncertainty is often ambiguous. Moreover, in the geosciences, repeated experiments are rarely possible due to the non-stationary nature of the Earth system, making it difficult to operationally define aleatoric uncertainty. For this reason, we do not explicitly distinguish between the two. Instead, we adopt a practical Bayesian perspective, interpreting all uncertainties as expressions of rational belief informed by available evidence. 

To discuss uncertainty quantification in deep learning, we first introduce some notation. Deep learning networks define a mapping from input vector $x$ to output vector $z$ via a neural network:
\begin{equation}
z=f(x,w),
\label{eq:map}
\end{equation}
where $f(x,w)$ denotes the neural network function parameterized by the weight vector $w$, which is learned from a dataset of input-output pairs. The function $f$ comprises nested nonlinear transformations, enabling the model to fit highly complex relationships in the data. The flexibility of neural networks stems from the high dimensionality of $w$, which often includes thousands of parameters (weights and biases).

Several sources contribute to the uncertainty in the output z. From Eq. (\ref{eq:map}), we can identify direct contributions from uncertainty in the input vector $x$, uncertainty in the weight vector $w$, and uncertainty in the functional form of $f(..)$ because the neural network will not be perfect in mapping input to the output vector. Additionally, since $w$ is inferred from training and testing data, uncertainties in these datasets must also be considered, even if not explicitly shown in Eq. (\ref{eq:map}). Another critical source of uncertainty arises when a new input $x$ lies outside the range of the training data. However, because this extrapolation uncertainty lacks a principled treatment (see the Discussion), we focus on the first four sources in this study. Once quantified, these uncertainties should be combined to yield a total predictive uncertainty described by a probability density function (pdf). Yet, no existing practical method consistently incorporates all these sources into a single pdf.

To better explain the underlying difficulty and the incompleteness of existing methods for uncertainty quantification, we need to refer to Bayes Theorem. This theorem states that the pdf of a random variable $A$ given knowledge of a variable $B$, $p(A|B)$, can be expressed as the pdf of random variable $A$ without knowing $B$, so $p(A)$, and the likelihood $l(B|A)$, as:
\begin{equation}
p(A|B) = \frac{l(B|A)}{p(B)}p(A).
\label{eq:Bayes}
\end{equation}
We call $p(A)$ the prior pdf of $A$, and $p(A|B)$ the posterior pdf of $A$ given $B$. The pdf of $B$ in the denominator is a normalization factor because $B$ is not a random variable, but has a given value. $l(B|A)$ represents the likelihood that $A$ gives rise to $B$, which is a function of random variable $A$. It determines how likely it is to find value $B$ for each value of random variable $A$. For future reference, we also use the theorem in its conditional form:
\begin{equation}
p(A|B,C) = \frac{l(B|A,C)}{p(B|C)}p(A|C),
\label{eq:Bayescond}
\end{equation}
which is just Bayes Theorem with all the factors conditioned on another set of variables $C$.

In the following, we will discuss a few common methods to provide uncertainty estimates in machine learning, and provide a first critical discussion of their basic assumptions. As we will see, among the main sources of uncertainty mentioned above, most attention has been given to the uncertainty in the weight vector $w$ and the uncertainty in the trained neural network $f(..)$. 

\subsection{Bayesian Deep Learning}
In theory, the uncertainty of $w$ in an neural network can be derived directly from Bayes Theorem, which is the basis of Bayesian Deep Learning, \cite{MacKay1992, Neal1996, Andrieu2003}. From Eq. (\ref{eq:Bayes}), if we choose $A=w$ and $B$ as the training data, we know that the posterior uncertainty in $w$ given a training dataset is determined by the prior uncertainty in $w$, and a likelihood that indicates how likely the training data are if the weight vector is equal to a certain $w$. 

Since the function $f(..)$ in deep learning is highly nonlinear, this likelihood, and hence the uncertainty pdf for the weight vector $w$, will have an extremely complex shape (e.g., many local minima and maxima). For low-dimensional systems, one can use methods such as Markov-Chain Monte Carlo (MCMC), Langevin sampling, and Hybrid or Hamiltonian MC to accurately describe the posterior pdf \cite{Andrieu2003, VanLeeuwen2015b}, but these methods do not scale well for high-dimensional problems, as is often the case in deep learning applications. To alleviate this issue, approximations to the posterior pdf have been proposed, such as the Gaussian or Laplace approximation, or variational Bayes (approximated pdf by a finite set of pdfs of a certain shape, e.g., a Gaussian mixture). These methods need the computation of the Hessian, the second derivative of the loss function, which can be computationally expensive. Apart from the crude Gaussian approximation, all these methods remain very expensive in high-dimensional problems, which pose difficulties in practice \cite{Blundell2015}. 

\subsection{Bagging and Bootstrapping}
A popular machine-learning method for uncertainty quantification related to the weights $w$ is Bootstrapping or Bagging \cite{Breiman1996}, which uses an ensemble of weight vectors to generate an ensemble of outputs. In this method, the weight vector ensemble can be generated in several ways, e.g., by randomly drawing initial weights and by randomly varying the order in which training batches are used during training of the neural network, or by giving each neural network a subset of the training data. When a new input vector is presented to the neural network, each trained weight vector in the resulting ensemble is used to generate one output value and interpreted to have the same probability. This, then, results in an ensemble of output values that is suggested to represent its uncertainty pdf. Unfortunately, while this method gives a spread in the output values, this spread does not represent a proper uncertainty estimate. The reason is that during training, the ensemble of initial weight vectors is transformed into an ensemble of trained weight vectors. Because this transformation is nonlinear, the pdf of the equally probable initial weight vectors is deformed into a pdf with a complicated shape determined by the likelihood of each weight vector to produce the training data (corresponding to the $l(B|A)$ factor above). Hence, the trained weight vectors are not equally likely anymore, and neither are the output values based on the neural network's that use these weights. The simple treatment of equally likely output values, as assumed in Bagging, is inappropriate. 

This issue with the likelihood of the weight vectors is more troublesome than might be anticipated. In principle, the likelihood value of each trained weight vector, also called the importance weight, can be properly calculated and included to evaluate the ensemble of weight vectors. However, in practice, one weight vector typically fits the training data significantly better than the others, resulting in a much higher importance weight. This dominance effectively collapses the ensemble to a single neural network, eliminating the diversity needed for meaningful uncertainty estimation and yielding an unrealistically low (often near-zero) uncertainty. While this challenge is acknowledged in Bayesian Deep Learning, where methods have been developed to construct ensembles of approximately equally likely weight vectors, it is overlooked in Bagging and related approaches, essentially making the uncertainty quantification useless. We return to this critical issue in detail in Section 3.3.

\subsection{Monte-Carlo Dropout}
Another popular method for uncertainty quantification related to the weights $w$ is Monte-Carlo (MC) Dropout \cite[]{Gal2016}. In conventional dropout, a method used to avoid overfitting, the weight of each neuron is set to zero during the training phase with a probability $P$. After training, all weights are used, but their trained value is multiplied by $P$ to generate one neural network. In Monte-Carlo Dropout, the same dropout procedure used during training is also used during prediction. In other words, the trained weights are also put randomly to zero with probability $P$ during prediction. This prediction procedure is performed several times to generate an ensemble of weight vectors, leading to an ensemble of output values. This output ensemble is then suggested to provide an uncertainty estimate. 

The issue with MC Dropout is that during training one realization of a trained weight vector is created. This weight vector is randomly perturbed by dropout to generate an ensemble of perturbed trained weight vectors. What should be the next step is to importance-weigh each of those perturbed weight vectors by their likelihood value, as discussed above for Bagging, but that step is omitted. Hence, similar to Bagging, the uncertainty quantification is inappropriate. Furthermore, it fails a fundamental requirement of Bayesian inference: the posterior uncertainty should decrease as more data becomes available. This behavior has been shown to be absent in MC Dropout  \cite[and references in these papers]{Osband2016, Verdoja2021}.

\subsection{Estimating uncertainty distribution parameters directly}
The above discussion focused on uncertainties in the weight vector $w$. A second category of uncertainty-quantification approaches tries to estimate the pdf of output uncertainty directly. Instead of predicting output $z$, the neural network predicts:
\begin{equation}
p_{\alpha}(z)=f_{\alpha} (x,w_{\alpha} )
\end{equation}
where $p_{\alpha}(z)$ denotes the uncertainty pdf of output $z$ with parameters $\alpha$, and the subscript $\alpha$ on the weight vector $w$ and on $f$ denotes that the target of estimation are the distribution parameters $\alpha$, not $z$ itself as in Eq (\ref{eq:map}). The difference from the methods above is that the uncertainty pdf in $z$ is not determined via an ensemble of weight vectors $w$, but instead, one weight vector is trained to provide the pdf parameters directly. Often, a Gaussian assumption is used, in which the mean and variance are estimated; however, other standard distributions can also be used \cite[]{Nix1994, Wang2016}.

\subsection{Evidential deep learning}
Evidential deep learning for deep regression is an extension of the above methodology in which not only the parameters of the pdf of $z$ are learned, but also the uncertainty in those parameters, e.g. \cite{Amini2020}. Specifically, a Normal-Inverse-Gamma distribution is used to describe the uncertainty in the mean and variance parameters. In this way, part of the epistemic or structural uncertainty is addressed. 

\subsection{Quantile regression}
While the two methods discussed above assume a certain shape for the uncertainty pdf, typically Gaussian, in quantile regression the pdf is allowed to have any shape, at least in principle. The neural network predicts:
\begin{equation}
P_z=f_P (x,w_P )
\end{equation}
where $P_z$ denotes predefined quantiles of the uncertainty pdf of output $z$, and the subscript $P$ on the weight vector $w$ and on $f$ denotes that the target of estimation iare these quantiles \cite[]{Wang2016, Pfreundschuh2018, Sonderby2020}. Quantile regression works well for a low-dimensional output space but becomes impractical in high-dimensional problems, because the number of output variables required to describe histograms increases dramatically with the dimension. To avoid the latter issue, this method is often used to estimate the marginal uncertainty pdfs instead of the full joint pdf.

\subsection{Deep Ensembles}
However, the methods in the second and third categories fail to account for the uncertainty in the weight vector $w_p$. This omission is partly corrected in so-called Deep Ensembles \cite[]{Lakshminarayanan2017}. Deep Ensembles generates an ensemble of pdf parameters, or, alternatively, an ensemble of histograms, by training multiple networks, each initialized with different untrained weight vectors and using distinct training batch orders. The resulting ensemble of pdf parameters is further averaged to represent the pdf of $z$, or the histograms themselves are averaged. Similar to Bagging and MC dropout, the simple average of the ensemble overlooks the fact that the trained weight vectors $w_p$, and thus the output ensemble members, are not equally likely.  

As mentioned above, the literature has concentrated on quantifying uncertainty in the weight vector $w$ or tried to estimate the uncertainty pdf of $z$ directly. None of the machine-learning methods discussed above has considered the uncertainty in the training and testing data, or in the new input data. While uncertainties in the training, testing, and new input data might be small in some fields, they can be significant in the geosciences, where real-world data, such as direct measurements, model outputs, or reanalysis data, often contain substantial errors that cannot be ignored. 

\subsection{Conformal prediction}
A completely different category is conformal prediction \cite{Balasubramanian2014}. It is based on the concept of dissimilarity, or nonconformity, of a new prediction compared to previous input data and their predictions from the neural network, so that the prediction interval size is larger for new “unusual” instances and smaller for “typical” instances. First, the neural network model is trained on a dataset. Then, a nonconformity score is defined by the user, which measures the extent to which a data point deviates from the model’s predictions. For instance, one can use the difference between the predicted value and the true value as a score. Given new input data, conformal prediction uses the nonconformity scores from the testing data to calculate a prediction interval. The idea is to find the smallest set or interval such that, with a predefined confidence level (like 90\%), the true value falls inside. Hence, the method produces a confidence interval, not an uncertainty estimate as Bayesian methods do, and it ignores uncertainty in the weight vector. Furthermore, in the geosciences, the training, testing, and new input data all contain uncertainties, and especially including uncertainty in output training data is not straightforward with this methodology. In fact, one typically relies on Bayesian methods to do this, resulting in an interesting, but perhaps ad hoc mixture of Bayesian and significance testing methodologies.

\subsection{Recent reviews on uncertainty quantification}
To conclude, while many methods have been developed for quantifying uncertainty in the output of a deep-learning neural network, they are either incomplete or statistically inappropriate. Recent articles by \cite{Abdar2021,Gawlikowski2023}, amd \cite{He2025} provide comprehensive overviews of existing techniques, but they lack critical discussions of the statistical properties of these methodologies. As a result, understanding the true meaning of uncertainty estimates from these approaches remains challenging. The review by \cite{Cheng2023} discusses the connection between data assimilation, machine learning, and uncertainty quantification, but the uncertainty quantification methods mentioned above are quoted without a critical review. 

\subsection{Our contribution}
In this paper, we present a novel approach that accounts for all sources of uncertainty. We derive an expression for the exact pdf and provide an efficient methodology for practical computation. In Section 2, we outline how to systematically incorporate all uncertainty sources. In Section 3, we introduce a practical approach to calculate the relevant uncertainties and combine them effectively. Specifically, we explain how proposal densities, a standard method in statistics, can be used to address the likelihood issue associated with trained weight vectors $w$. In Section 4, we demonstrate the methodology on a simple toy problem and on an application to predict cloud process rates and offer a physical interpretation of the resulting uncertainty. Finally, we conclude with a discussion in the last section.

\section{Uncertainty propagation in Deep Learning}

The uncertainty in the output z can be described by the pdf  $p(z|x,\theta_{tr},\theta_{te} )$, which is the pdf of $z$ given an input vector $x$ and training and testing data $\theta_{tr}$ and $\theta_{te}$, respectively. The training and testing data consist of input and output pairs $(X,Z)$. The first step in deriving $p(z|x,\theta_{tr},\theta_{te} )$ is to incorporate the uncertainty in the input vector $x$. Following ideas from data assimilation \cite[]{VanLeeuwen2015b, VanLeeuwen2020, Evensen2022}, we can consider $x$ as a random draw from a distribution $p(x)$, centered around a true input vector. This true input vector is not known, but, as we will show later, this is not a concern. Let us consider all possible true input vectors that would lead to us drawing $x$. We denote these possible true inputs by $x^t$. We now use the identity:
\begin{equation}
p(A|C) = \int p(A,B|C)p(B)\;dB
\end{equation}
to incorporate the uncertainty in the new input vector x as follows:
\begin{eqnarray}
p(z|x,\theta_{tr},\theta_{te} )  & = &  \int p(z,x^t |x,\theta_{tr},\theta_{te})\;dx^t \nonumber \\
& = &  \int p(z|x^t,x,\theta_{tr},\theta_{te})p(x^t |x,\theta_{tr},\theta_{te})\;dx^t \nonumber \\
& = & \int p(z|x^t,\theta_{tr},\theta_{te})p(x^t |x)\;dx^t,
\end{eqnarray}
where for the second equality, the identity
\begin{equation}
p(A,B|C) = p(A|B,C)p(B|C)
\label{eq:condition}
\end{equation}
is used. For the third equality, we used that, if the true input was $x^t$, then x does not provide extra information on $z$. Furthermore, since input vectors do not depend on the training or testing data, we have $p(x^t|x,\theta_{tr},\theta_{te} )= p(x^t |x)$. This pdf describes the probability of each possible true input $x^t$ given that we have input sample $x$. This pdf is exactly the uncertainty pdf of $x$, as that uncertainty pdf describes the probability of the true input by definition.

The second step is to incorporate the uncertainty in the training and testing data. We can consider the training and testing data pairs as a random draw from a distribution $p(\theta_{tr},\theta_{te})$ centered around the true data $(\theta_{tr}^t,\theta_{te}^t)$. Similar to the way of how the uncertainty in the new input vector $x$ is handled, let us consider all possible true training and testing data $(\theta_{tr}^t,\theta_{te}^t)$ that would lead us to draw $(\theta_{tr},\theta_{te})$. This allows us to write:
\begin{eqnarray}
p(z|x,\theta_{tr},\theta_{te})  &= & \int p(z|x^t,\theta_{tr},\theta_{te})p(x^t |x)\;dx^t \nonumber \\
& = & \int \! \!\int \!\!\int p(z,\theta_{tr}^t,\theta_{te}^t|x^t,\theta_{tr},\theta_{te})p(x^t |x)\;dx^t d\theta_{tr}^t,d\theta_{te}^t \nonumber \\
& = & \int \! \!\int \!\!\int p(z|x^t,\theta_{tr}^t,\theta_{te}^t)p(\theta_{tr}^t |\theta_{tr})p(\theta_{te}^t|\theta_{te})p(x^t |x)\;dx^t d\theta_{tr}^t,d\theta_{te}^t,
\end{eqnarray}
where we used the identity in Eq. (\ref{eq:condition}) and the same reasoning as for the new input $x$. We also used that the uncertainties in training and testing data are independent.

We now incorporate the uncertainty of weight vector that defines the neural network. We can write
\begin{eqnarray}
p(z|x,\theta_{tr},\theta_{te})  &= &  \int \! \!\int \!\!\int p(z|x^t,\theta_{tr}^t,\theta_{te}^t)p(\theta_{tr}^t|\theta_{tr})p(\theta_{te}^t|\theta_{te})p(x^t |x)\;dx^t d\theta_{tr}^t,d\theta_{te}^t \nonumber \\
& = &  \int \! \!\int \!\!\int \!\! \int p(z,w|x^t,\theta_{tr}^t,\theta_{te}^t)p(\theta_{tr}^t|\theta_{tr})p(\theta_{te}^t|\theta_{te})p(x^t |x)\;dx^t d\theta_{tr}^t,d\theta_{te}^t dw.
\label{eq:w1}
\end{eqnarray}
We now evaluate further
\begin{equation}
p(z,w|x^t,\theta_{tr}^t,\theta_{te}^t) = p(z,|w,x^t,\theta_{tr}^t,\theta_{te}^t)p(w|x^t,\theta_{tr}^t,\theta_{te}^t) = p(z|w,x^t,\theta_{te}^t)p(w|\theta_{tr}^t)
\label{eq:w2}
\end{equation}
where we used that if the weight vector is given, the training data do not provide extra information on $z$. Furthermore, the weight vector $w$ by definition is only dependent on the training data and does not depend on the testing data and $x^t$.

Putting Eq. (\ref{eq:w2}) in Eq. (\ref{eq:w1}), we find:
\begin{equation}
p(z|x,\theta_{tr},\theta_{te}) =  \int \! \!\int \!\!\int \!\! \int p(z|w,x^t,\theta_{te}^t)p(w|\theta_{tr}^t)p(\theta_{tr}^t|\theta_{tr})p(\theta_{te}^t|\theta_{te})p(x^t |x)\;dx^t d\theta_{tr}^t,d\theta_{te}^t dw.
\label{eq:final}
\end{equation}
The right-hand side of Eq. (\ref{eq:final}) highlights the sources of the uncertainty in $z$, namely 
\begin{itemize}
\item[1.] Uncertainty in the new input, captured by $p(x_t |x)$, 
\item[2.] Uncertainty in the data pairs in the training and testing sets, described by $p(\theta_{tr}^t|\theta_{tr})$ and $p(\theta_{te}^t|\theta_{te})$, 
\item[3.] Uncertainty in the weights, represented by $p(w|\theta_{tr}^t)$, 
\item[4.] Intrinsic uncertainty in the neural network, reflecting model imperfection, given by  $p(z|w,x^t,\theta_{te}^t)$.  
\end{itemize}

This equation can be interpreted as the {\it fundamental equation of machine learning}, as it provides a complete probabilistic characterization of a new output sample. Such characterization represents the ultimate goal of machine learning: not merely to predict a value, but to quantify the uncertainty in that prediction. It shifts the paradigm away from producing a single “best” estimate—an often ill-defined and insufficient approach for real-world applications—towards fully probabilistic descriptions, which are essential for informed decision-making and scientific rigor.

The only uncertainty not included is uncertainty in the neural network architecture. That uncertainty can be included in a manner similar to uncertainty in the weights, but a complication is that the dimension of the weight vector will change with architecture. This is not a fundamental problem, but requires careful consideration of probability spaces beyond the scope of this paper. 

The next section discusses practical ways to estimate the four pdfs. 

\section{A practical method for uncertainty quantification}
\subsection{Source 1: Uncertainty in the new input}
The influence of uncertainty in the new input vector $x$ on the output $z$ is the most straightforward to implement. One can draw $N_x$ samples from the probability density function $p(x_t |x)$,which describes the uncertainty around the observed input vector $x$. Each of these sampled input vectors is then propagated through the neural network to generate corresponding samples of the output $z$. This Monte Carlo approach provides an empirical approximation of the distribution of $z$ induced by the uncertainty in $x$. Mathematically, this means that we draw the samples from the uncertainty pdf, $p(x^t|x)$, as: 
\begin{equation}
p(x^t|x) = \frac{1}{N_x} \sum_{i=1}^{N_x} \delta(x^t - x_i^t),
\label{eq:x-samples}
\end{equation}
where $x_i^t$ are the drawn samples centered around $x$. Using this expression, we can rewrite Eq. (\ref{eq:final}) as:
\begin{equation}
p(z|x,\theta_{tr},\theta_{te}) =  \frac{1}{N_x}\sum_{i=1}^{N_x}\int \!\!\int \!\! \int p(z|w,x_i^t,\theta_{te}^t)p(w|\theta_{tr}^t)p(\theta_{tr}^t|\theta_{tr})p(\theta_{te}^t|\theta_{te})\;d\theta_{tr}^t,d\theta_{te}^t dw.
\label{eq:uncertainx}
\end{equation}
%

\subsection{Source 2: Uncertainty in the training and testing data}
The uncertainty in the training and testing data can be treated in the same way as the uncertainty in the new input vector $x$. Similar to Eq. (\ref{eq:x-samples}), we draw samples from $p(\theta_{tr}^t|\theta_{tr})$ and $p(\theta_{te}^t|\theta_{te})$ and write:
\begin{equation}
p(\theta_{tr}^t|\theta_{tr}) =\frac{1}{N_1} \sum_{j1=1}^{N_1}\delta(\theta_{tr}^t-\theta_{tr,j1}^t)
\end{equation}
and similar for $p(\theta_{te}^t|\theta_{te})$. This leads to:
\begin{equation}
p(z|x,\theta_{tr},\theta_{te}) =  \frac{1}{N_x}\sum_{i=1}^{N_x} \frac{1}{N_1} \sum_{j1=1}^{N_{X1}}\frac{1}{N_2} \sum_{j2=1}^{N_{X2}}\int p(z|w,x_i^t,\theta_{te,j2}^t)p(w|\theta_{tr,j1}^t) \; dw,
\label{eq:uncertaintr}
\end{equation}
where $N_{X1}$ and$N_{X2}$ are the number of samples drawn from the uncertainty pdf of the input-output pairs in the training and testing datasets, respectively. Note that we have $N_{X1}$ samples for each training input-output pair, and $N_{X2}$ samples for each testing input-output pair.

\subsection{Source 3: Uncertainty in the weights}
\subsubsection{Basic derivation}
We now turn to the estimation of the weight distribution conditioned on the training data in Eq. (\ref{eq:uncertaintr}), namely $p(w|\theta_{tr,j1}^t)$. As discussed in the Introduction, this pdf is inherently complex due to the highly nonlinear relationship between the neural network weights and the training data. Consequently, the exact form of this distribution is unknown, and direct sampling from it is not feasible.

Instead, we adopt a pragmatic approach inspired by standard practices in machine learning. In typical training routines, the weights are obtained through a minimization procedure (e.g., stochastic gradient descent), given a fixed neural network architecture and a predefined optimization algorithm. Under these conditions, the final weight vector is determined primarily by two stochastic components: the initialization of the weights and the order in which the training data batches are processed. By varying these components, we can generate multiple plausible realizations of the trained weights. These realizations serve as approximate samples from the unknown distribution $p(w|\theta_{tr,j1}^t)$, enabling a Monte Carlo approximation of the pdf. However, as explained below, considerable effort is needed to obtain a practical algorithm.

Let us first introduce the random batch order in our formalism and derive an expression for $p(w|\theta_{tr,j1}^t)$ that takes this uncertainty into account. By denoting the batch order $b$, we can write:
\begin{equation}
p(w|\theta_{tr,j1}^t) = \int p(w,b|\theta_{tr,j1}^t)\;db = \int p(w|b,\theta_{tr,j1}^t) p(b|\theta_{tr,j1}^t) \; db = \int p(w|b,\theta_{tr,j1}^t) p(b) \; db,
\label{eq:weights1}
\end{equation}
where we used the fact that the batch order is independent of the training data. By further employing Bayes Theorem in the form of Eq. (\ref{eq:Bayescond}), and writing the input-output pair of $\theta^t_{tr,j1}$ as $(X^t_{tr,j1},Z^t_{tr,j1})$, we find:
\begin{equation}
p(w|b,X^t_{tr,j1},Z^t_{tr,j1}) = \frac{l(Z^t_{tr,j1}| w,b,X^t_{tr,j1})}{p(Z^t_{tr,j1}| b,X^t_{tr,j1})}p(w|b,X^t_{tr,j1}).
\label{eq:weights2}
\end{equation}
We can evaluate $p(w|b,X^t_{tr,j1})= p(w)$, since the input data $X^t_{tr,j1}$ and the batch order $b$ alone have no information on the weights $w$. Therefore, the conditional pdf is equal to the prior distribution over the weights. 

The likelihood $l(Z^t_{tr,j1}| w,b,X^t_{tr,j1})$ quantifies how well a particular weight vector $w$, when applied to the input data $X^t_{tr,j1}$ in the batch order $b$, reproduces the corresponding output data $Z^t_{tr,j1}$ of the training dataset. As explained in the Introduction, the denominator, $p(Z^t_{tr,j1}| b,X^t_{tr,j1})$ serves as a normalization constant that ensures the posterior is a valid pdf. Since we are primarily interested in drawing samples from the posterior rather than evaluating it explicitly, this term does not need to be computed in practice.

Inserting Eq. (\ref{eq:weights2}) into Eq. (\ref{eq:weights1}) we find:
\begin{equation}
p(w|\theta_{tr,j1}^t)  =  \int  \frac{l(Z^t_{tr,j1}| w,b,X^t_{tr,j1})}{p(Z^t_{tr,j1}| b,X^t_{tr,j1})}p(w)p(b)\; db
\label{eq:weights3}
\end{equation}
In the next section, we focus on determining the likelihood $l(Z^t_{tr,j1}| w,b,X^t_{tr,j1})$ and show how to perform the integral.  

\subsubsection{The likelihood and the integral}
To determine the likelihood, we use the relationship between the training pairs and the weight vectors as specified by the neural network:
\begin{equation}
Z_{tr,j1}^t = f(X^t_{tr,j1},w,b) + \epsilon(X_{tr,j1}^t),
\label{eq:model}
\end{equation}
where  $\epsilon(X_{tr,j1}^t)$ is the difference between the predicted $f(X_{tr,j1}^t,w,b)$  and the output of training $Z_{tr,j1}^t$. Because we already incorporated the uncertainty in the input and output training data above in Section 3.2, we can interpret this difference as the uncertainty in the neural network function itself. Rewriting this relation as $f(X^t_{tr,j1},w,b)=Z_{tr,j1}^t - \epsilon(X_{tr,j1}^t)$, and since $Z_{tr,j1}^t$ is given, the uncertainty distribution of $f(X^t_{tr,j1},w,b)$ is equal to that of $\epsilon(X_{tr,j1}^t)$ shifted by the constant vector $Z_{tr,j1}^t$.

As an example, assume the uncertainty in $\epsilon(X_{tr,j1}^t)$ to be Gaussian, with zero mean and error variance $\sigma_f^2$ for each element $m$ of vector $\epsilon(X_{tr,j1}^t)$. We then find the likelihood:
\begin{equation}
l(Z^t_{tr,j1}| w,b,X^t_{tr,j1}) = A \exp\left [ - \frac{1}{2} \sum_{m=1}^{N_{tr}} \frac{\left(Z_{m,tr,j1}^t - f(X_{m,tr,j1},w,b)\right)^2}{\sigma_f^2} \right],
\label{eq:likelihood}
\end{equation}
where $A$ is a normalization factor, $N_{tr}$ is the total number of the input-output pairs in the training dataset. The term under the exponent can be recognized as a typical machine-learning loss function.

We now turn to the calculation of the integral in Eq. (\ref{eq:weights3}). We can draw initial weight vector samples $w_k$ from $p(w)$ and batch order samples $b_k$ from $p(b)$, which leads to:
\begin{equation}
p(w|\theta_{tr,j1}^t)  = \frac{1}{N_w} \sum_{k=1}^{N_w}  \frac{l(Z^t_{tr,j1}| w_k,b_k,X^t_{tr,j1})}{p(Z^t_{tr,j1}| b_k,X^t_{tr,j1})}\delta(w-w_k),
\label{eq:weights4}
\end{equation}
where $N_w$ is the total number of drawn weight vector samples. This expression shows that the posterior pdf of the weight vectors is represented by the initial weight vectors $w_k$, each with a so-called importance weight: 
\begin{equation}
\alpha_{j1,k} =  \frac{1}{N_w}  \frac{l(Z^t_{tr,j1}| w_k,b_k,X^t_{tr,j1})}{p(Z^t_{tr,j1}| b_k,X^t_{tr,j1})}
\label{eq:importance}
\end{equation}
This implies that any posterior moment of the weight vector $w$ (e.g., the mean or covariance) must be computed using importance weights $\alpha_{j1,k}$. However, in practice, the values of $\alpha_{j1,k}$ tend to vary dramatically between samples and, typically, one of them takes a value close to 1, while the others are nearly zero. This is because the likelihood contains an exponential term, as shown in Eq. (\ref{eq:likelihood}). The sum in the brackets varies considerably with the weight vector $w_k$, and the exponent of that sum will enhance that variation substantially. As a result, the posterior distribution becomes sharply peaked, and only a single weight vector significantly contributes to the estimate of $p(w|X_{tr,j1}^t,Z_{tr,j1}^t)$. The effective ensemble size collapses to one, eliminating any representation of uncertainty. This issue is known as filter degeneracy in the particle filter literature \cite[]{Doucet2001, VanLeeuwen2019, Evensen2022}. It renders the use of Eqs. (\ref{eq:weights4}) and (\ref{eq:importance}) impractical, as they fail to maintain a diverse and informative ensemble of weight vectors.

\subsubsection{The proposal density}
Unlike Eq. (\ref{eq:weights4}),  where samples of the weight vector $w$ are drawn from the prior distribution $p(w)$, we can instead draw samples from a distribution of trained weights. This approach is mathematically formalized by introducing a {\it proposal density} $q(w|X_{tr,j1}^t,Z_{tr,j1}^t )$ which explicitly depends on the training data  \cite[]{Doucet2001, VanLeeuwen2010, Ades2012}. To incorporate this, we multiply and divide the prior pdf $p(w)$ by the proposal density $q(w|X_{tr,j1}^t,Z_{tr,j1}^t )$. This procedure will allow us to perform importance sampling using samples drawn from this proposal rather than from the $p(w)$. Using this idea in Eq. (\ref{eq:weights3}), we find:
\begin{eqnarray}
p(w|\theta_{tr,j1}^t)& = & \int  \frac{l(Z^t_{tr,j1}| w,b,X^t_{tr,j1})}{p(Z^t_{tr,j1}| b,X^t_{tr,j1})}p(w)p(b)\; db \nonumber \\
& = & \int  \frac{l(Z^t_{tr,j1}| w,b,X^t_{tr,j1})}{p(Z^t_{tr,j1}| b,X^t_{tr,j1})}\frac{p(w)}{q(w|b,X_{tr,j1}^t,Z_{tr,j1}^t )}q(w|b,X_{tr,j1}^t,Z_{tr,j1}^t )p(b)\; db \nonumber \\
& = &  \frac{1}{N_w} \sum_{k=1}^{N_w}  \frac{l(Z^t_{tr,j1}| w_{j1,k},b_k,X^t_{tr,j1})}{p(Z^t_{tr,j1}| b_k,X^t_{tr,j1})}\frac{p(w_{j1,k})}{q(w_{j1,k}|b_k,X_{tr,j1}^t,Z_{tr,j1}^t )}\delta(w-w_{j1,k}),
\label{eq:weights5}
\end{eqnarray}
where the weight vectors $w_{j1,k}$ are now drawn from the proposal density $q(w|b_k,X_{tr,j1}^t,Z_{tr,j1}^t )$. The conditioning on the training data means that the weight vectors have been trained on the training data. The details on how they are trained are described in the proposal density. 

A standard choice in machine learning is to draw the weight vectors by minimizing a loss function, which is the same as maximizing the likelihood. This minimization is combined with a methodology to avoid overfitting. It is not easy to write down the proposal density in this case. No matter these details, from Eq. (\ref{eq:weights5}) we see that $p(w|\theta_{tr,j1}^t)$ is now represented by a set of weights $w_{j1,k}$, each with importance weight given as:
\begin{equation}
\beta_{j1,k} = \frac{1}{N_w}   \frac{l(Z^t_{tr,j1}| w_{j1,k},b_k,X^t_{tr,j1})}{p(Z^t_{tr,j1}| b_k,X^t_{tr,j1})}\frac{p(w_{j1,k})}{q(w_{j1,k}|X_{tr,j1}^t,Z_{tr,j1}^t )}.
\end{equation}
The $\beta_{j1,k}$ consist of three factors that depend on the value of the weight vector, the prior pdf, the proposal density, and the likelihood. The prior $p(w)$ is typically a uniform distribution around zero, i.e., the probability $p(w_{j1,k})$ is the same for any drawn weight vector $w_{j1,k}$. The proposal density $q(w|b_k,X_{tr,j1}^t,Z_{tr,j1}^t )$ does depend on the trained weight vector. To see this, denote a prior weight vector sample as $u_k$ and the corresponding trained weight vector as $w_{j1,k}$. The proposal density is related to the prior density via the standard transformation of variables formula from statistics, as:
\begin{equation}
q(w_{j1,k}|b_k,X_{tr,j1}^t,Z_{tr,j1}^t ) = \left| \frac{\partial w_{j1,k}}{\partial u_k}\right|_{b_k} p(u_k),
\label{eq:proposal}
\end{equation}
where $\left| \partial w_{j1,k}/\partial u_k\right|_{b_k}$ is the Jacobian of this transformation. This Jacobian is a complicated function and depends on $u_k$ and $b_k$. Although different prior samples $u_k$ will have the same value of $p(u_k)$, the Jacobians are different for different $u_k$ and $b_k$, and, consequently, the proposal densities of the trained weights will be different from each other. The likelihood values of the trained weights will also be different because some trained weight vectors will fit the training data better than others. This means that the importance weights $\beta_{j1,k}$ will be different for each weight vector. It is well known from the particle filtering literature that if the training data set is large, the differences in the $\beta_{j1,k}$ will be huge, again leading to one importance weight having a value close to 1, while all others have values very close to zero. 

Since Bagging, Deep Ensembles, and MC dropout use the trained weights in this way, they are of little value for providing information on $p(w|\theta^t_{tr,j1})$. However, we can construct a proposal density that does not have this problem, as explained below.

\subsubsection{A new proposal density}
To overcome the filter degeneracy issue, we design the proposal density $q(w|X_{tr,j1}^t,Z_{tr,j1}^t )$ such that all importance weights $\beta_{j1,k}$ are equal, or at least nearly so. One intuitive approach is to enforce a fixed likelihood value across all sampled weights. Specifically, we define a target likelihood value $l_0$ and seek intermediate weight vectors $\tilde{w}_{j1,k}$ that satisfy the identity:
\begin{equation}
l(Z^t_{tr,j1}| \tilde{w}_{j1,k},b_k,X^t_{tr,j1}) = l_0.
\end{equation}
To ensure that the $\beta_{j1,k}$ are equal, all values $q(\tilde{w}|X_{tr,j1}^t,Z_{tr,j1}^t )$  need to be equal as well. According to Eq. (\ref{eq:proposal}) this implies that the Jacobian of the transformation from the untrained weights $u_k$ to the trained weights $\tilde{w}_{j1,k}$ must not depend on $k$. However, variations in $u_k$ that change $\tilde{w}_{j1,k}$ such that $l(Z^t_{tr,j1}| \tilde{w}_{j1,k},b_k,X^t_{tr,j1})$  increases or decreases are not allowed, because the likelihood has to be fixed at $l_0$. This means that the matrix $(\partial \tilde{w}_{j1,k}/\partial u_k)$ is not full rank, and its determinant vanishes:
\begin{equation}
 \left|\frac{ \partial \tilde{w}_{j1,k}}{\partial u_k}\right|=0
 \end{equation}
resulting in a proposal density $q(\tilde{w}|X_{tr,j1}^t,Z_{tr,j1}^t )$ that is not well defined. As a result, while this method is conceptually appealing, it is not mathematically viable for practical use.

To avoid this issue, we define the final weight vectors in the proposal density by adding a small perturbation of size $\delta$ to each intermediate weight vector $\tilde{w}_{j1,k}$, as:
\begin{equation}
w_{j1,k} = \tilde{w}_{j1,k} + \delta r_{j1,k}(u_k),
\end{equation}
where $r_{j1,k}(u_k)$ is a function of the untrained weight vector $u_k$. The Jacobian can now be evaluated via a Taylor-series expansion around $\left| \partial w_{j1,k}/\partial u_k\right|$, and, since $\delta$ is small, we only need to retain up to the first term in the expansion, leading to:
\begin{equation}
\left| \frac{\partial w_{j1,k}}{\partial u_k}\right| = \left| \frac{\partial \tilde{w}_{j1,k}}{\partial u_k} + \delta \frac{\partial r_{j1,k}}{\partial u_k}\right| 
\approx \left| \frac{\partial \tilde{w}_{j1,k}}{\partial u_k} \right| + \delta Tr\left[Adj\left(\frac{\partial \tilde{w}_{j1,k}}{\partial u_k}\right)\left(\frac{\partial r_{j1,k}}{\partial u_k} \right)\right] + O\left(\delta^2\right),
\end{equation}
where $Tr$ denotes the trace, and $Adj$ denotes the adjugate. If we now choose $(\partial r_{j1,k}/\partial u_k)$  as the pseudoinverse $(\partial r_{j1,k}/\partial u_k)=Adj\left(\frac{\partial \tilde{w}_{j1,k}}{\partial u_k}\right)^{\dagger}$ we find:
\begin{equation}
\left| \frac{\partial w_{j1,k}}{\partial u_k}\right| =   \delta n+ O\left(\delta^2\right),
\end{equation}
where $n = \dim(w_{j1,k})$, and we used that $\left| \partial \tilde{w}_{j1,k}/\partial u_k\right|=0$. This shows that, up to order $\delta$, the Jacobian is a constant, independent of the weight vector $w_{j1,k}$. This, then, means that the proposal density $q(w_{j1,k}|X_{tr,j1}^t,Z_{tr,j1}^t )$ in Eq. (\ref{eq:proposal}) has the same value for each weight vector $w_{j1,k}$.

However, by adding a small perturbation to the trained weight vectors, the likelihood is no longer exactly equal to $l_0$. To assess the impact of this perturbation, we perform a Taylor series expansion on the likelihood around the trained weights. This yields:
\begin{equation}
l(Z^t_{tr,j1}| w_{j1,k},b_k,X^t_{tr,j1}) = l_0+ \delta r_{j1,k} \nabla_{ w_{j1,k}} l + O(\delta^2),
\end{equation}
in which $\delta r_{j1,k}$ is the pertubation. Since the weights $\tilde{w}_{j1,k}$ have been obtained through training, the likelihood is already near its maximum, implying that the gradient $\nabla_{ w_{j1,k}} l $ is small. By choosing the perturbation magnitude $\delta$ to be sufficiently small, the first-order term becomes negligible, and the perturbed likelihood remains very close to $l_0$. This approximation restores the practical viability of the method.

The target value for the likelihood $l_0$ is determined as follows. First, we train the neural network with the unperturbed training data, resulting in a weight vector denoted by $w_0$. Next, we use the $N_{X1}$ perturbed versions of the training data and pass each of them through the neural network using this fixed weight vector $w_0$. This process yields $N_{X1}$ loss function values $J_{j1} (Z_{tr,j1}^t, X_{tr,j1}^t,w_0 )$. We average these as  $J_0  =1/N_{X1}   \sum_{j1=1}^{N_{X1}}J_{j1} (Z_{tr,j1}^t, X_{tr,j1}^t,w_0 )$ and take as target likelihood value $l_0=\exp(-J_0)$. By defining $l_0$ this way, we obtain a representative likelihood value for trained weight vectors. Consequently, the set of trained weight vectors obtained from the perturbed training datasets can be interpreted as an unbiased sample from the posterior distribution of the weights, $p(w|X_{tr,j}^t,Z_{tr,j1}^t)$.

With the above construction, where each weight vector $w_{j1,k} \approx \tilde{w}_{j1,k}$, we have ensured that the proposal densities $q(w_{j1,k}|X_{tr,j1}^t,Z_{tr,j1}^t )$ are identical across all samples. Additionally, the corresponding likelihoods $l(Z^t_{tr,j1}| w_{j1,k},b_k,X^t_{tr,j1})$ are the same for each weight vector. As a result, all importance weights $\beta_{j1,k}$ are equal, implying that each weight vector $w_{j1,k}$ is equally probable in the posterior distribution $p(w|X_{tr,j}^t,Z_{tr,j1}^t)$. 

With these equally probable samples from $p(w_{j1,k}|X_{tr,j1}^t,Z_{tr,j1}^t )$, Eq. (\ref{eq:uncertaintr}) can be rewritten as:
\begin{equation}
p(z|x,\theta_{tr},\theta_{te}) =  \frac{1}{N_x}\sum_{i=1}^{N_x} \frac{1}{N_{X1}} \sum_{j1=1}^{N_{X1}}\frac{1}{N_{X2}} \sum_{j2=1}^{N_{X2}} \frac{1}{N_w} \sum_{k=1}^{N_w} p(z|w_{j1,k},x_i^t,X_{te,j2}^t,Z_{te,j2}^t).
\label{eq:total3}
\end{equation}
%

\subsection{Uncertainty in the trained neural network}
As shown in Eq. (\ref{eq:total3}), the final step is to estimate  $p(z|w_{j1,k},x_i^t,X_{te,j2}^t,Z_{te,j2}^t)$, which captures the uncertainty due to the neural network's imperfection. This imperfection is evaluated using the testing data $(X_{te,j2}^t,Z_{te,j2}^t)$ for each index $j2$. Since the evaluation is based on the statistical relation described by the input-output pairs of testing data, we rewrite 
\begin{equation}
p(z|w_{j1,k},x_i^t,X_{te,j2}^t,Z_{te,j2}^t)=p_{\theta_{te,j2,i}} (z|w_{j1,k},x_i^t),
\label{eq:weightpdf}
\end{equation}
i.e., the pdf of $z$ given the input $x_i^t$ and weight vector $w_{j1,k}$, determined using the input-output pairs in the testing data $j2$. The extra index $i$ in the testing data denotes that we only use testing data for which the input data is close to $x_i^t$. Specifically, we calculate the pdf of the output testing data $Z_{te,j2}$ for which the input testing data is within a region around the new input, so $X_{tr,j2}^t \in D_{x_i^t }$. This pdf can be approximated by known functions (e.g., a Gaussian or a Gaussian mixture), by a histogram, by the set of samples $Z_{te,j2,m}$, or by new samples generated from $Z_{te,j2,m}$ (e.g., via a so-called diffusion process \cite[]{Sohl2015}). 

The size of $D_{x_i^t }$ is governed by the uncertainty in the input testing data. In our experience, setting the radius of $D_{x_i^t }$  between 0.5 and 2 times the uncertainty standard deviation in the input testing data yields robust and consistent output pdf estimates.

\subsection{The resulting algorithm and practical considerations}
Using Eq. (\ref{eq:weightpdf}) into Eq. (\ref{eq:total3}), we have a complete uncertainty quantification methodology that can be written as:
\begin{equation}
p(z|x,\theta_{tr},\theta_{te}) =  \frac{1}{N_x}\sum_{i=1}^{N_x} \frac{1}{N_{X1}} \sum_{j1=1}^{N_{X1}}\frac{1}{N_{X2}} \sum_{j2=1}^{N_{X2}} \frac{1}{N_w} \sum_{k=1}^{N_w} p_{\theta_{te,j2,i}} (z|w_{j1,k},x_i^t).
\end{equation}
This equation shows that we average the pdfs $p_{\theta_{te,j2,i}} (z|w_{j1,k},x_i^t)$ over the input samples $i$, the trained weight vector samples $k$, the training data samples $j1$, and the testing data samples $j2$, to find the full uncertainty pdf of the output of the deep learning procedure. The following provides the set of steps needed in this calculation:

\begin{itemize}
\item[1)] {\bf Initial training}: Train a neural network using the original, unperturbed training data. Apply any standard technique to avoid overfitting, but do not use dropout. Store the resulting weight vector as $w_0$.
\item[2)] {\bf Generate perturbed training data}: Create an ensemble of $N_{X1}$  training datasets drawn from the uncertainty distributions of the input and output training datasets. This results in $N_{X1}$ perturbed versions of the full training dataset. 
\item[3)] {\bf Determine the target likelihood value $l_0$}: Use the perturbed input training datasets from step 2 to calculate their loss function values using the corresponding perturbed training output dataset and the fixed weight vector $w_0$ from step 1. Average the $N_{X1}$ loss function values and store the result in $J_0$. Calculate the target likelihood value $l_0=\exp(-J_0)$. 
\item[4)] {\bf Determine an ensemble of weight vectors}: For each perturbed training dataset, train $N_w$ neural networks with different initial untrained weight vectors and batch orders. This leads to $N_{X1} \times N_w$ different neural networks. These networks should all be trained such that their loss function value is close to the target loss function value $J_0$. This process results in $N_{X1}\times N_w$ trained weight vectors.
\item[5)] {\bf Generate perturbed new input data}: Create $N_x$  perturbed new input values drawn from the uncertainty distribution of the new input value. This results in $N_x$ perturbed versions of the new input value.
\item[6)] {\bf Generate perturbed testing data}: Create $N_{X2}$  perturbed testing datasets drawn from the uncertainty distribution of the testing data. This results in $N_{X2}$ perturbed versions of the testing data. 
\item[7] {\bf Generate samples for the pdf}: For each perturbed new input value from step 5, find all the perturbed testing input data from step 6 that are close to it. Pass the selected testing input data through the $N_{X1}\times N_w$ neural networks from step 4 and subtract the result from their corresponding testing output data. This results in a large number of output samples for each perturbed new input data.
\item[8)] {\bf Determine final pdf}: Calculate histograms for each perturbed new input data based on the samples calculated in step 7. Average the histograms to find the final uncertainty pdf for output $z$.
\end{itemize}

\subsection{Practical tips}
\begin{itemize}
\item[1] Ensemble sizes. We tested our algorithm on several test cases, including a real-world application described below. In our experience, the predictive pdf converged for $N_x=N_{tr}=N_{te}=N_w=20$ in all applications. However, we expect these numbers to be application dependent. 
\item[2] Accuracy of likelihood calculations. The likelihood is calculated from the loss function by multiplying the loss function by minus one, and then taking the exponent. When a large amount of training data is used, the loss function values will be very large, and taking the exponent of minus that large number can lead to underflow issues. Since we are only interested in relative likelihood values, we can   subtract the same constant value from each loss function before taking the exponent. In our algorithm we used the $J_0$ value. 
\item[3] Simplifications for Gaussian output uncertainties. The above algorithm can be simplified significantly if $l(Z_{tr,j1}^t |w_{j1,k},b_k,X_{tr,j1}^t)$ is assumed to be Gaussian as in Eq. (\ref{eq:likelihood}), and the uncertainty in the output training data $Z_{tr}$ is also Gaussian distributed. In that case, we do not need to introduce perturbed output training data $Z_{tr}^t$ in Step 2. Instead, we can work directly with $p(w|X_{tr,j1}^t,Z_{tr})$. Then the likelihood of the weight vectors becomes (see Appendix A):
\begin{equation}
l(Z^t_{tr,j1}| w,b,X^t_{tr,j1}) = A \exp\left [ - \frac{1}{2} \sum_{m=1}^{N_{tr}} \frac{\left(Z_{tr,m}^t - f(X_{tr,j1,m},w,b)\right)^2}{\sigma_f^2+\sigma_{z,m}^2} \right],	
\end{equation}
in which $\sigma_{z,m}^2$ is the uncertainty variance in the output training data $Z_{tr,m}$. All steps and equations from Eq (\ref{eq:likelihood}) to Eq. (\ref{eq:total3}) remain unchanged, with the only modification being that the perturbed output training vector $Z_{tr,j1}^t$ is replaced everywhere with the original, unperturbed output vector $Z_{tr}$. This is a situation often encountered in the geosciences. 
\end{itemize}

\section{Comparison of methodologies on a simple regression problem}
To demonstrate how the uncertainty characterized by our method differs from existing approaches, we compare results on a simple regression problem with the Bagging ensemble method by \cite{Breiman1996} and the pdf-based methodology proposed by \cite{Wang2016}. Our implementation of the pdf-based method is quantile regression.

\begin{figure}[htp]
\centerline{\includegraphics[width=0.9\textwidth]{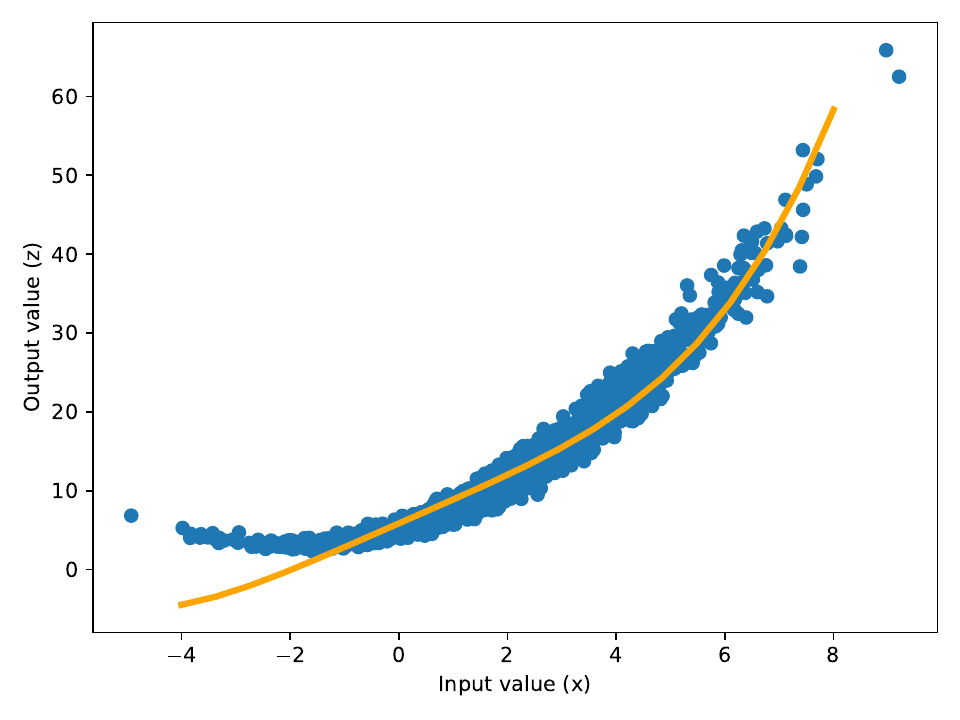}}
\caption{Training and testing data (blue dots) and example model (orange line).}
\label{fig:SimpleModelSamples}
\end{figure}

The training and testing data sets are generated from the following system
\begin{equation}
z=ax^2  + bx + c
\label{eq:TrueModel}
\end{equation}
in which $a=0.5$, $b=2.0$, and $c=5$. A total of 2,000 input–output samples are generated, with inputs drawn from a normal distribution with a mean of 2 and a standard deviation of 2. Both the input and output variables are perturbed by additive Gaussian noise with a standard deviation of 0.3. The full dataset is shown in Fig. \ref{fig:SimpleModelSamples}. The 2000 input-output samples are divided into 1600 training and 400 testing samples.All samples are displayed in Fig. \ref{fig:SimpleModelSamples}. 

Based on the relationship observed in Fig. \ref{fig:SimpleModelSamples}, or guided by prior knowledge of the underlying system, one might select the following neural network configuration to represent the system behavior:
\begin{equation}
z = dx^4 + ex + f,
\end{equation}
In this example, the neural network weight vector is $w = (d,e,f)^T$. 

The orange curve in Fig. \ref{fig:SimpleModelSamples} illustrates the machine-learned solution based on the unperturbed training data. While the model fits well for positive input values, it underperforms for inputs smaller than -2.  This outcome illustrates a common issue: even when a model fits the training data adequately, its structural constraints can restrict its ability to generalize across the full input domain.

\begin{figure}[htp]
\centerline{\includegraphics[width=0.99\textwidth]{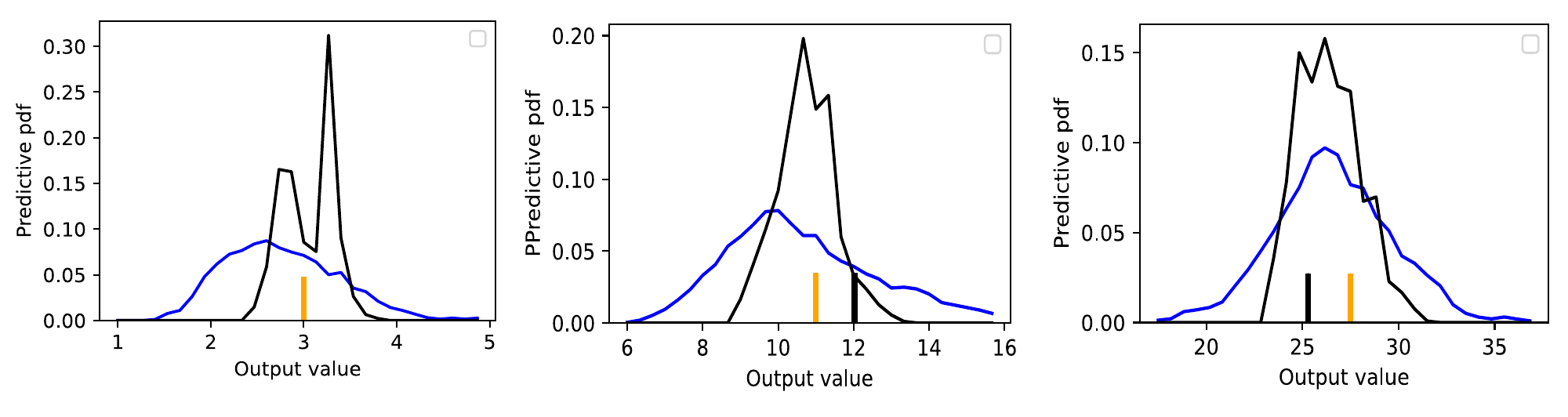}}
\caption{Posterior pdfs derived from Bagging (black vertical line), quantile regression (black curve), and new methodology (blue line), for input values -2 (left), 2 (middle), and 5(right). The pdfs can be compared with the samples in Fig. \ref{fig:SimpleModelSamples}. The orange vertical line is the prediction using the true model. (Note that this not the true prediction because the input value has uncertainty.) In the left panel the Bagging vertical lines fall around -0.07, outside the plot range. Bagging shows a degenerate pdf, and the quantile regression pdf is too narrow because uncertainties in training, testing, and new input data are ignored.}
\label{fig:SimpleModel}
\end{figure}

Figure \ref{fig:SimpleModel} presents the results for three new input values, specifically $x = -2$, $x = 2$, and $x = 5$. As a reference, using these new inputs in conjunction with the true equation (\ref{eq:TrueModel}), we compute the corresponding 'true' output values. These values are shown as orange vertical bars in Figure 2, under the assumption that the input values are known precisely and have no uncertainty.

The results from the Bagging method are shown by the black bars in Figure \ref{fig:SimpleModel}. In the left panel, the Bagging vertical lines fall around -0.07, outside the plot range. These bars represent the output values from an ensemble of 20 neural networks. Due to the high similarity among the ensemble members, the individual predictions are nearly indistinguishable and collectively appear as a single thick black bar. As discussed in earlier sections, this outcome reflects a fundamental limitation of the Bagging approach. Variations in initial weights or the order of training batches have only a minimal effect on the final trained weights in this simple regression problem. Consequently, all networks converge to similar solutions, leading to a significant underestimation of the associated uncertainty. This limitation is concerning, because in more complex real-world applications, Bagging may similarly fail to capture meaningful uncertainty, potentially giving a false sense of confidence in the predictions. Another fundamental issue is the bias displayed by the Bagging procedure. The values follow the orange curve in Fig. \ref{fig:SimpleModelSamples} quite closely, so a bias in the neural network becomes a bias in Bagging. The reason the new method and quantile regression do not have this bias problem is that they use the testing samples to define the pdfs, not only the weight values. MC dropout would show the same issue as bagging, while Deep Ensembles, based on quantile regression pdfs, would not have this strong bias. 

Quantile regression estimates the linear relationship between inputs and outputs at specified quantiles. Given the nonlinear nature of the system, as illustrated in Fig. \ref{fig:SimpleModelSamples}, a linear fit over all output values for each quantile is clearly inappropriate.  To address this, the quantile regression is restricted to an interval around the new test input value.  For the results shown in Fig. \ref{fig:SimpleModel}, we used an interval spanning ten standard deviations of the input error. Alternative interval sizes were tested and yielded comparable outcomes. As shown in Fig. \ref{fig:SimpleModel}, quantile regression (black curves) results in uncertainty estimates that are too narrow. This is expected because uncertainties in training, testing, and new input data are neglected. 

Compared to the existing methods, the new method provides an uncertainty estimate that is broader than that of quantile regression. To see that it is more realistic, one can add typical input uncertainties of 0.3 to the new input values, and study the output spread in Fig. \ref{fig:SimpleModelSamples}. The proof that the new method provides the correct uncertainty estimate follows directly from its formulation: it takes all uncertainty sources properly into account. The only approximation is in the size of the ensembles used to characterize the uncertainties in the individual sources. Increasing these sizes by a factor 10 leads to the same visual pdfs as displayed in Fig. \ref{fig:SimpleModel}, strongly suggesting that the method has converged. 

We can conclude that none of the existing uncertainty quantification methods provides a proper uncertainty estimate in this simple system. In practical applications, hyperparameters in these existing methods are often tuned to produce visually plausible uncertainty intervals, but such tuning lacks generalizability and must be repeated for each new application. The proposed method avoids this limitation and provides consistent uncertainty estimates.

\section{Comparison of methodologies on predictions of cloud process rate}
In this section, we implement our methodology to quantify the uncertainty associated with autoconversion rates. Autoconversion is a fundamental microphysical process that governs the formation of raindrops through cloud droplet collision and coalescence. Due to its inherently stochastic and nonlinear nature, autoconversion is well-suited for machine learning based prediction and uncertainty quantification \cite[]{Chiu2021, Gettelman2021}.

\subsection{Description of the training and testing datasets}
In this study, we leverage the training and testing datasets used in \cite{Chiu2021}. The input variables are derived from aircraft cloud probe measurements collected during the Aerosol and Cloud Experiments in the Eastern North Atlantic (ACE-ENA) field campaign, operated by the Atmospheric Radiation Measurement (ARM) user facility. These include cloud water content ($q_c$), cloud droplet number concentration ($N_c$), drizzle water content ($q_r$), and drizzle drop number concentration ($N_r$), with respective uncertainties of 30\%, 50\%, 30\%, and 20\% (\cite[]{Glienke2019, Glienke2020, Mei2020}. The output variable, autoconversion rate, is calculated by applying the in-situ cloud droplet size distribution to the stochastic collection equation formulated as a two-moment bin model. The rate spans a wide dynamic range, from $7.7\times10^{-24}$ to $2.8\times10^{-6}\; kg\ m^3 s^{-1}$, capturing diverse stages of cloud evolution and supporting robust training. The uncertainty in the autoconversion rate is approximately 53\%, based on the analytical autoconversion equation (6) from \cite{Chiu2021}. The full dataset, consisting of approximately 10,612,000 samples, is partitioned into training, validation, and testing subsets, with a 60-20-20\% data split. To reduce dynamic range and satisfy the assumption of additive Gaussian noise, a base-10 logarithmic transformation is applied to all input and output variables.
 
\subsection{Training procedure}
Building on the neural network architecture used by \cite{Chiu2021}, we implement a feedforward network comprising six fully connected hidden layers, each containing 16 nodes. Following the procedure outlined in Section 3.5, we first trained the network using the unperturbed training dataset, with randomly initialized weights and a shuffled batch order. To mitigate overfitting, we applied the early stopping criterion proposed by Prechelt (1998), halting training when the validation loss no longer decreased with additional epochs. Given the high variability of the validation loss across epochs, we smoothed the loss curve using a moving average over a 50-epoch window. We then identified the epoch at which the smoothed loss stabilized, and used the corresponding model weights as the final trained weights, $w_0$.

Following the initial training with unperturbed data, we then trained separate neural networks for each of the 20 perturbed training datasets $X_{tr,j1},Z_{tr,j1}$, using 20 different sets of perturbed initial weights and batch orders. This yielded $N_w =20$ models per perturbed dataset. Training was stopped when the loss function for the training data reached $J_0$. As a result, we obtained a total of 400 neural networks (20 networks for each of the 20 perturbed datasets) to quantify uncertainty in both the model weights and the training data.

\begin{figure}[htp]
\centerline{\includegraphics[width=0.95\textwidth]{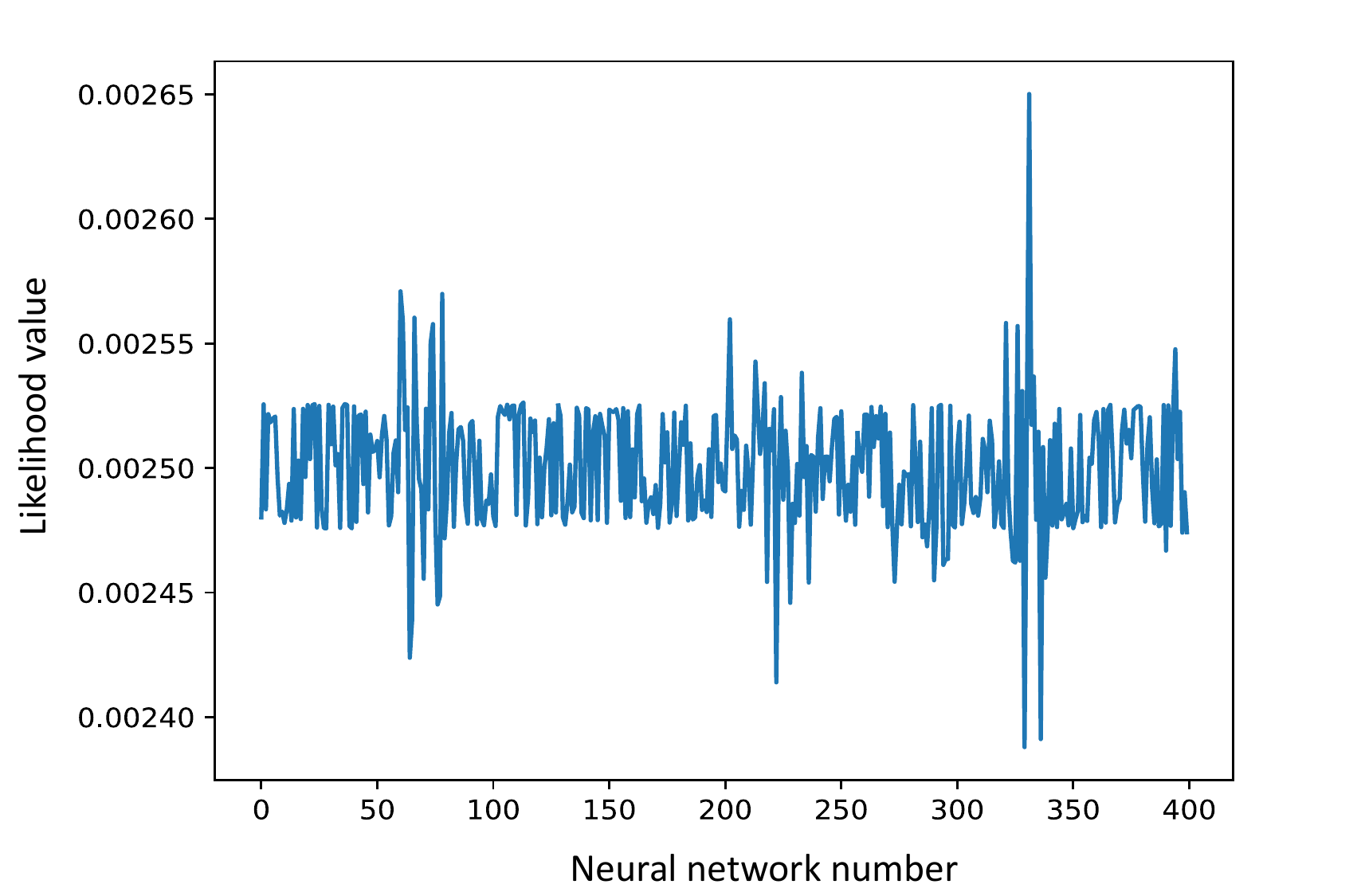}}
\caption{The normalized likelihood values (i.e., the importance weights) for the $400$ neural networks.}
\label{fig:LikelihoodValues}
\end{figure}

In practice, exact convergence to $J_0$ is complicated by the fact that many machine learning frameworks report loss only at the end of each epoch. To address this, we identified the two epochs where the loss values bracketed $J_0$, and then applied a bisection method to approximate a loss value within 0.01 of  $J_0$. This threshold ensures that the likelihood values are sufficiently close to $l_0 = \exp(-J_0)$, leading to approximately uniform importance weights across the ensemble. As shown in Fig. \ref{fig:LikelihoodValues}, most importance weights are close to the expected value of $1/400=0.0025$, which corresponds to a perfect equal-weight ensemble. A small number of networks did not converge precisely, which is expected due to the high dimensionality of the optimization space. For these networks, training was stopped at slightly less precise loss values, resulting in importance weights deviating by no more than 6\%, which is acceptable for our purposes. 
 
\subsection{Results}
To evaluate our methodology on new input data, we selected 50 distinct input vectors spanning a wide range of values. Each input vector was perturbed 100 times using random noise sampled from the input uncertainty distributions described in Section 4.1. This resulted in 40,000 output samples per input vector ($N_x=100$, $N_w=20$, $N_{X1}=20$). For each input, we computed the predictive uncertainty distribution using the testing data, as described in Eq. (29), and averaged the resulting histograms.

As mentioned in Section 3.4, the region ($D_{x_i^t }$) around each perturbed new input $x_i^t$, must be defined to compute the local empirical distribution from the test data. We used all output testing data that had corresponding input testing data close to $x_i^t$ within $0.2$ (in $\log10$ space), for each of the four variables in the input vector. We experimented with thresholds ranging from 0.1 to 0.5 and found that this choice had no significant visual impact on the resulting distributions.

The results for all $50$ input vectors are displayed in Appendix B. To highlight the key findings, Fig.  \ref{fig:Full_pdfs} shows the total uncertainty pdfs as a function of the $\log10$ autoconversion rate for four representative input vectors. In each panel, the blue curve is the total uncertainty pdf from Eq. (30), and the red bar is the point prediction from the baseline neural network (i.e., using $w_0$) without uncertainty quantification. The black bars show the predictions from Bagging, obtained by creating an ensemble of 20 networks with different initial weights and batch orders.

\begin{figure}[htp]
\centerline{\includegraphics[width=0.8\textwidth]{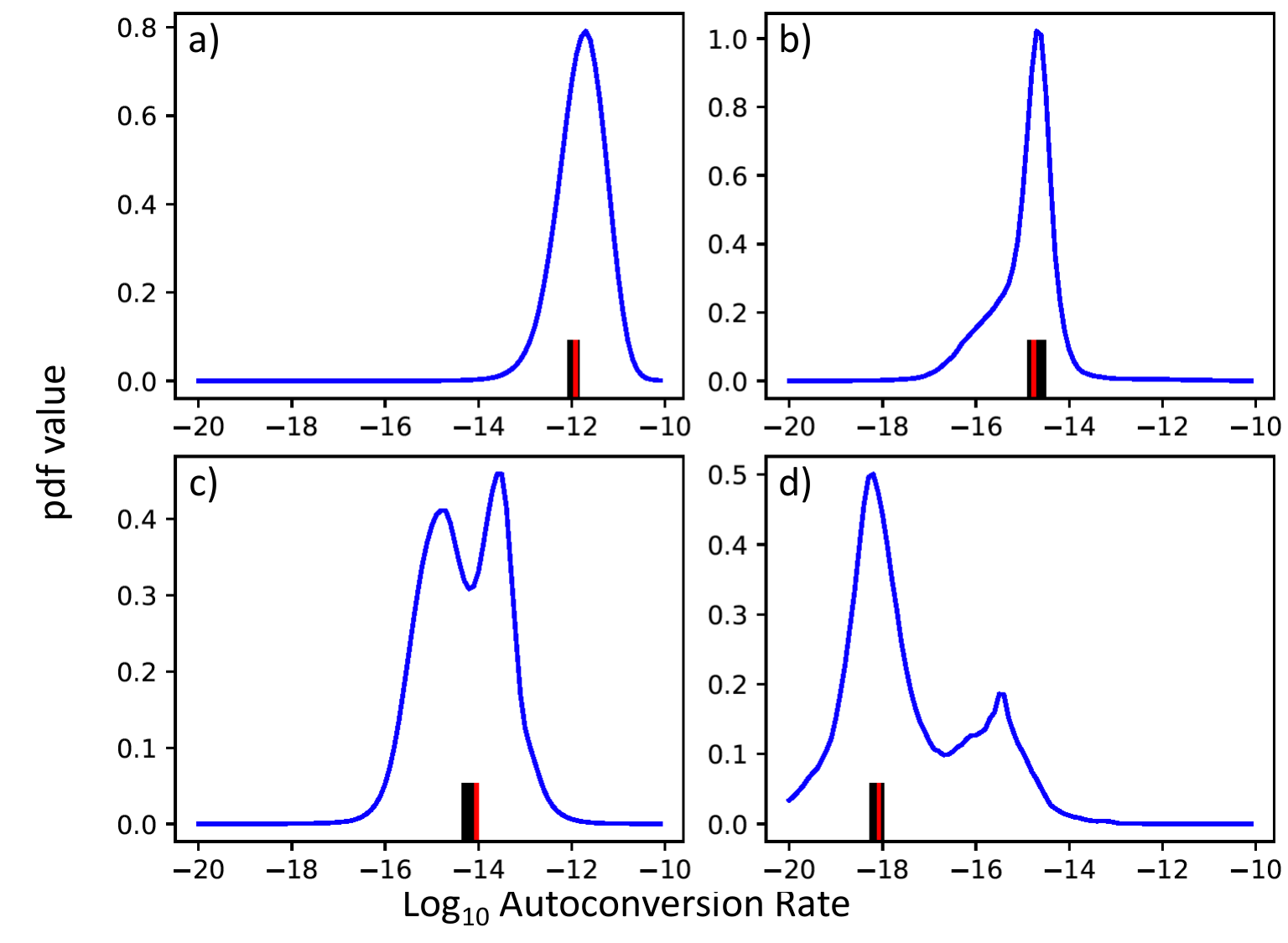}}
\caption{Examples of total uncertainty pdf in the output autoconversion rate for four input vectors. The blue curves are the total uncertainty pdfs; the red bar is the autoconversion rate calculated by the baseline neural network ($w_0$) without uncertainty quantification; black bars are the Bagging output samples. Note the wide variety of shapes of the blue uncertainty pdfs and the small spread in the black bars, demonstrating the inadequacy of the Bagging approach to represent uncertainty.}
\label{fig:Full_pdfs}
\end{figure}

Figures  \ref{fig:Full_pdfs}(a) and (b) show that the predictive pdfs span roughly an order of magnitude in autoconversion rate, which is not surprising given the relatively large measurement uncertainty in the input variables (ranging between 20\% and 50\%). Although the pdf in Fig. \ref{fig:Full_pdfs}(a) can be approximated by a Gaussian distribution, a non-negligible skewness is observed in Fig. \ref{fig:Full_pdfs}(b). The pdfs in Fig.  \ref{fig:Full_pdfs}(c) and (d) are broader, showing large uncertainty and bimodality. 

To disentangle the contributions of different uncertainty sources, we first isolated the effect of neural network weight uncertainty. This was done by disabling all other uncertainty sources in Eq. (30) and computing outputs using only the ensemble of weight vectors. The resulting distributions were very narrow and closely resemble the Bagging ensemble outputs (black bars), indicating that weight uncertainty is not a major contributor.

\begin{figure}[htp]
\centerline{\includegraphics[width=0.8\textwidth]{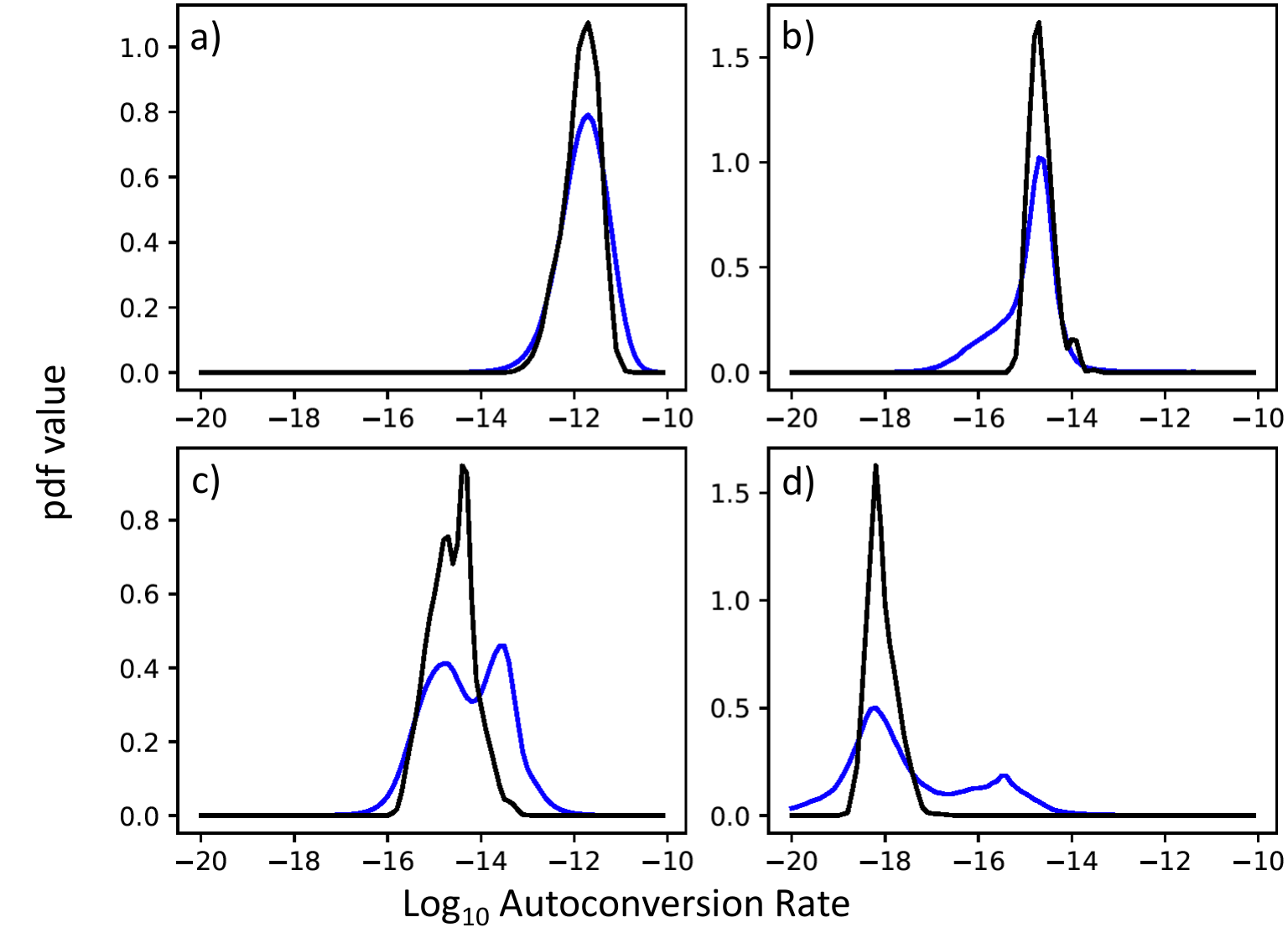}}
\caption{Same as Fig. \ref{fig:Full_pdfs}, showing full uncertainty pdfs (blue curves) and the contribution from input uncertainty and model uncertainty (black curves).}
\label{fig:Full_Input}
\end{figure}

Next, we assessed the combined impact of input uncertainty and model uncertainty from the testing data. This involved evaluating the corresponding terms in Eq. (30) while ignoring the other uncertainties. These two sources contributed similar amounts of variability and together accounted for most of the spread in the predictive distribution, as shown by the black curves in Fig. \ref{fig:Full_Input}, which peak at similar values to the full pdfs (blue curves). However, the combined input and model uncertainty distributions were typically unimodal and symmetric (in $\log_{10}$ space), failing to reproduce the skewness or bimodality observed in the full predictive distributions. This suggests that the observed skewness and bimodality are primarily caused by uncertainty in the training and testing datasets.

To further investigate the skewness and bimodality in the predictive pdfs, we analyzed all 50 cases in Appendix B. We found that positive skewness occurred predominantly at high autoconversion values (above $10-12$), while strong negative skewness was associated with intermediate values ($14-17$). Bimodal distributions were mainly observed at low autoconversion values (around $-18$). Attempts to relate these features to specific patterns in the 4D input space were inconclusive, and even 3D visualizations of input subsets failed to reveal consistent structure.
 
Returning to Fig. \ref{fig:Full_pdfs}, we observe the limitations of Bagging. The resulting ensemble predictions (black bars) exhibit little spread and fail to capture features such as bimodality, leading to significant underestimation of uncertainty. In fact, the shortcomings of Bagging extend beyond this. As noted in Section 3.3 (following Eq. 21), ensemble methods like Bagging, Deep Ensembles, and MC Dropout implicitly assume equal likelihood for all network realizations. However, in practice, the importance weights of these networks, derived from their likelihood values, vary exponentially.

To illustrate this further, we normalized all likelihoods by the largest value. This resulted in one network receiving an importance weight of 1, while the next highest weight was $\exp(-824)$, effectively zero. All remaining networks contributed negligibly. Thus, only the highest-likelihood network meaningfully affects the predictive distribution, rendering the ensemble effectively a single-model prediction. This outcome stands in stark contrast to our method, which produces near-uniform importance weights (see Fig. \ref{fig:LikelihoodValues}). Accordingly, if standard ensemble-based methods were applied in a statistically consistent manner, they would collapse to a single model and fail to provide meaningful uncertainty quantification.

\section{Conclusions and discussion}
Quantifying uncertainty in deep learning results is essential for their scientific application and for achieving broader acceptance of machine learning methodologies in the geosciences. In this study, we reviewed the existing literature on uncertainty quantification and highlighted common shortcuts, approximations, and sources of uncertainty that are often overlooked in current approaches. Given the limitations and inconsistencies in these methods, we developed a comprehensive and systematic framework for quantifying output uncertainty that accounts for all major sources:
\begin{itemize}
\item[1.] uncertainty in the input variables,
\item[2.] uncertainty in the training and testing data,
\item[3.] uncertainty in the neural network weight vectors (arising from variations in initial weights and batch order), and
\item[4.] uncertainty inherent to the neural network itself, as captured by its behavior on the testing dataset.
\end{itemize}

Our methodology is general and makes no assumptions about the relative importance of these uncertainty sources. It provides a scientifically rigorous foundation for machine learning applications and is expected to support the adoption of such methods in geoscientific research

We demonstrated the practical implementation of this approach using a simple regression example and a real-world example: predicting cloud autoconversion rates with a deep learning model trained on a two-moment bin formulation of the stochastic collection equation and aircraft cloud probe measurements from the ACE-ENA field campaign, operated by the Atmospheric Radiation Measurement (ARM) user facility. Our analysis revealed that the dominant source of predictive uncertainty in this case stems from the uncertainty in the training and testing datasets, contributing to variations in output of about an order of magnitude. The uncertainty from the new input data and from the neural network itself, as determined by the testing data, also made significant contributions, roughly half that magnitude. In contrast, uncertainty due to initial weight values and training batch order was found to be comparatively minor.

An important question is how our methodology scales with the size of the deep-learning problem. From a practical viewpoint, similar ensemble sizes were needed in the simple and the real-world example, suggesting a scaling independent of system size. However, our real-world example is relatively small, so our findings might not generalize. Nevertheless, we know from numerical weather prediction that ensemble sizes of the order of 20-100 are sufficient to quantify uncertainty in highly nonlinear systems with up to $10^{11}$ variables, see, e.g., \cite{Isaksen2010} and \cite{VanLeeuwen2019}. The reason a relatively small ensemble works is that the focus in the geosciences is on marginal pdfs of one or a few variables, and not the complete joint posterior pdf. As long as the objective remains to estimate (many) marginal pdfs, we see no fundamental reason why this conclusion would not extend to deep learning. This suggests that ensemble sizes of 20–100 should generally suffice, even for large-scale problems.

We further compared our results with widely used ensemble-based uncertainty quantification techniques, including Bagging, Deep Ensembles, and Monte Carlo (MC) Dropout. These methods, in their standard implementations, generate ensembles of model predictions but fail to account for the correct statistical weighting of each ensemble member. When proper importance weighting based on likelihood values is applied, only one member dominates the ensemble, rendering the rest negligible. This effectively collapses the ensemble to a single model, making these methods unsuitable for rigorous uncertainty quantification.

One could argue that existing ensemble-based methods, such as Bagging, Deep Ensembles, or MC Dropout and methods based on them, can be improved by tuning hyperparameters in the neural network to better capture output uncertainty. While it is true that carefully tuned networks can yield more realistic uncertainty estimates, this process is often computationally expensive and lacks generalizability across different applications or datasets. In contrast, a key advantage of our methodology is that it does not require any hyperparameter tuning. The only user-defined parameters involve the sizes of the ensembles, which is a requirement shared with ensemble-based methods. This makes our approach not only more principled but also more practical and broadly applicable.

Finally, we note that when new input data fall outside the domain of the training and testing sets, deep learning models can produce unreliable or even invalid predictions. This represents an additional source of uncertainty that is not explicitly included in our current framework because it cannot yet be quantified in a principled manner. However, two complementary strategies may help mitigate this issue. The first is to make the trained machine more robust by by expanding the training dataset to cover a broader region of the input space. Some methods have been proposed in this direction, e.g., using symmetries in training data \cite[]{Yu2021}. Our methodology automatically generates extra training and testing data, but does this in a principled way by perturbing these datasets within their uncertainty as part of the uncertainty quantification. The second methodology is to include extra prior knowledge in the model training process, for instance, extra constraints that enforce physical consistency. Bayes Theorem provides a systematic way to add such extra prior information. Since our methodology is rooted in Bayes’ Theorem, it provides a natural and rigorous foundation for including such constraints as priors, and doing so would be straightforward in practice.

In summary, our proposed framework offers a statistically consistent, comprehensive, and flexible solution for uncertainty quantification in deep learning applications. It is particularly well-suited to the geosciences, where complex models and measurement uncertainties demand a careful and principled approach.

\begin{appendix}\appheader
\section{Appendix. The uncertainty in the input training data}\label{appendixA}
In this appendix we provide an extended discussion of how one can treat uncertainty in input training data within the machine-learning likelihood.  Let the input training data be denoted by $X$, and the output training data by $Z$ for ease of notation. We also suppress batch ordering in the neural network function for clarity.  Following the main text, we introduce the true (but unknown) input data as a random variable $X^t$, and write:
\begin{equation}
p(Z|X,w) = \int p(Z,X^t|X,w)\;dX^t = \int p(Z|X^t,X,w)  p(X^t |X) \;dX^t = \int p(Z|X^t,w)  p(X^t |X)\;dX^t,
\label{eq:B1}
\end{equation}
where we used that the input data depends not on the weight vector, and that, with $X^t$ given, $X$ does not provide extra information on $Z$. The factor $p(X^t |X)$ is the pdf that describes the uncertainty in the input training data. 

For the other pdf, $p(Z|X^t,w)$, we use the fact that the weight vector $w$ is given, which means that the output training data $Z$ are assumed to be generated via a neural network with that weight vector, applied to the input training data $X^t$. In general, there will be uncertainty in the output training data and the neural network will not be perfect, so that we can write for each training pair $m$:
\begin{equation}
Z_m=f(X_m^t,w)+\epsilon_m (X_m^t ),
\end{equation}
where $f$ is the neural network and $\epsilon_m (X_m^t )$ denotes the uncertainty in the output training data $Z_m$ and in the neural network, which can potentailly depend on the input data value $X_m^t$. This equation shows that the uncertainty pdf of output training data given $X^t$ and $w$, so the pdf $p(Z|X^t,w)$, is the same as the pdf of $\epsilon(X^t )$ but shifted by the value $f(X^t,w)$. We used the same argument in Section 3.3, but note that we are now discussing uncertainty in training data, not the result of a new input value.

As an example, if we assume that the uncertainties in the output data and the neural network are independent Gaussian distributed, with variances $\sigma_{z,m}^2$ and $\sigma_f^2$, respectively, we can write:
\begin{equation}
p(Z_m |X_m^t,w)= p_{\epsilon} (\epsilon_m (X_m^t ))= A \exp \left[-\frac{1}{2}  \frac{(\epsilon_m^2 (X_m^t ))}{\sigma_f^2+\sigma_{z,m}^2 }\right]=A \exp \left[-\frac{1}{2}\frac{Z_m-f(X_m^t,w))^2}{\sigma_f^2+\sigma_{z,m}^2 }\right]	
\end{equation}
in which $A$ is a normalization factor and $\sigma_f^2+\sigma_{z,m}^2$ is the variance of the uncertainty variable $\epsilon_m (X_m^t )$. The independence assumption is not essential, and we refer to e.g. the data assimilation literature, such as \cite{Evensen2022}, on how to include dependencies in the uncertainties, but that is not the main point here. If we now also assume that the uncertainty in the input training data $p(X^t |X)$ is independent Gaussian, we find for the pdf for each training pair m, using Eq. (\ref{eq:B1}):
\begin{equation}
p(Z_m |X_m,w)=A \int \exp \left[-\frac{1}{2}\frac{Z_m-f(X_m^t,w))^2}{\sigma_f^2+\sigma_{z,m}^2} - \frac{1}{2}(X_m^t-X_m )^T C_m^(-1) (X_m^t-X_m )\right]dX_m^t,
\end{equation}
in which $A$ is another normalization factor, and the input $X_m$ is allowed to be a vector with uncertainty covariance $C_m$. (The independency is between input-output samples, not necessarily between the elements of an input sample.) Evaluating this integral is cumbersome because $f(X_m^t,w)$ is a highly nonlinear function. However, when the uncertainty in the input data is small, such that we can approximate: 
\begin{equation}
f(X_m^t,w) \approx f(X_m,w)+F_m (X_m^t-X_m ),
\end{equation}
in which $F_m$ is the local derivative of the nonlinear function f at input vector value $X_m$, we can evaluate the integral because its argument is now the product of two Gaussians in the integration variable $X_m^t$.  We then find:
\begin{equation}
p(Z_m |X_m,w)=A \int \exp \left[-\frac{1}{2}\frac{Z_m-f(X_m^t,w))^2}{\sigma_f^2+\sigma_{z,m}^2 + F_m C_m F_m^T} \right],
\label{eq:B3}
\end{equation}
Therefore, for all $N_X$ input data:
\begin{equation}
p(Z |X,w)=\prod_{m=1}^{N_X} p(Z_m |X_m,w) = A \int \exp \left[-\frac{1}{2}\sum_{m=1}^{N_X}\frac{Z_m-f(X_m^t,w))^2}{\sigma_f^2+\sigma_{z,m}^2 + F_m C_m F_m^T }\right],
\end{equation}
in which $A$ is yet another normalization constant. We conclude that, if the uncertainty in the input data is small, such that a linearization of the network around this input value is accurate, we recover the standard deep learning likelihood. However, the variance in that likelihood contains the sum of the output uncertainty, neural network uncertainty, and the input uncertainty transformed to output space via the linearized neural network (the $F_m C_m F_m^T $ term).  

When the uncertainties are not Gaussian or when the linearization of the neural network function is not accurate, analytical solutions quickly become impossible. In that case, we can use Monte-Carlo methods to evaluate the integral in Eq. (\ref{eq:B1})). The idea is to draw $N_x$ samples from $p(X_m^t |X_m )$. This means that we represent $p(X_m^t |X_m )$ by a set of delta functions, each centered on a sample $X_{m,j}^t$, as:
\begin{equation}
p(X_m^t |X_m )  =\frac{1}{N_X}\sum_{j=1}^{N_X} \delta (X_m^t-X_{m,j}^t ),
\end{equation}
where $\delta$ denotes the Dirac delta distribution. Substituting this in Eq. (\ref{eq:B1}), together with Eq. (B\ref{eq:B3}), we find for each training pair $m$:
\begin{equation}
p(Z_m |X_m,w)=A \frac{1}{N_X}\sum_{j=1}^{N_X} \exp \left[-\frac{1}{2}\frac{Z_m-f(X_{m,j}^t,w))^2}{\sigma_f^2+\sigma_{z,m}^2 } \right],
\end{equation}
such that 
\begin{equation}
p(Z|X,w)= A \frac{1}{N_X}\sum_{j=1}^{N_X} \exp \left[-\frac{1}{2}\sum_{m=1}^{N_X}\frac{Z_m-f(X_{m,j}^t,w))^2}{\sigma_f^2+\sigma_{z,m}^2 } \right], ,
\end{equation}
which is the formulation used in the main text.

\section{Appendix. Full results for the autoconversion case}\label{appendixB}
The following figures show the full uncertainty pdfs for all 50 new input values (blue curves), the uncertainty due to input uncertainty and the uncertainty of the network as determined from the testing data (black curves), the unperturbed autoconversion estimate (red bar) and the uncertainty estimate from the weight uncertainty only (black bars), but omitting the importance weights (see main text for discussion).
 
 \begin{figure}[htp]
 \centerline{\includegraphics[width=0.99\textwidth]{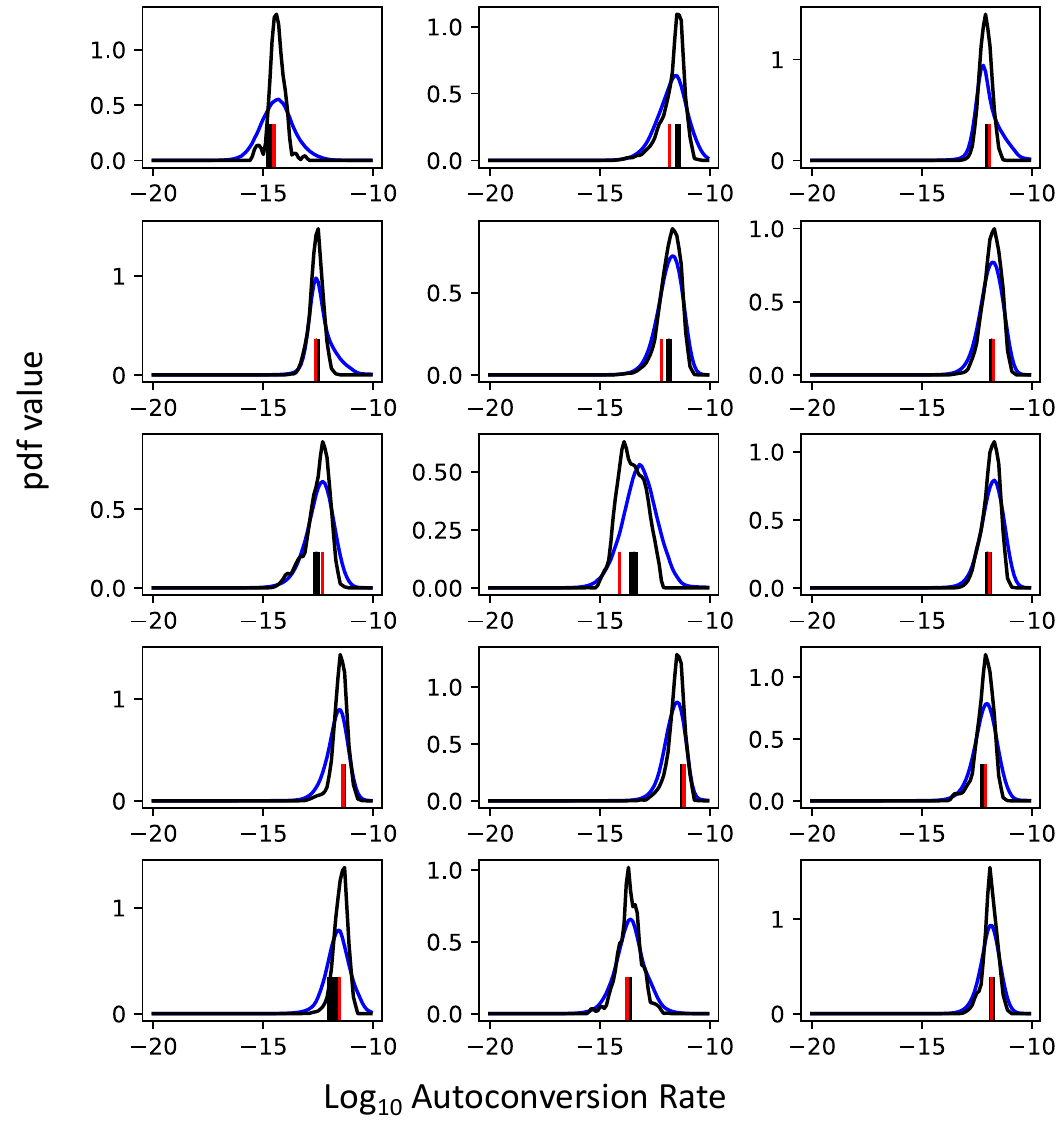}}
\caption{Examples of total uncertainty pdf in the output autoconversion rate for 50 input vectors. The blue curves are the total uncertainty pdfs,  black curves that are derived considering the uncertainty in the new input vector and the uncertainty in the neural network only. the red bar is the autoconversion rate calculated by traditional deep learning without uncertainty quantification. The black bars are the output samples generated using Bagging and used as a measure of uncertainty for the Bagging approach.}
\label{fig: B1}
\end{figure}

\begin{figure}[htp]
\centerline{\includegraphics[width=0.99\textwidth]{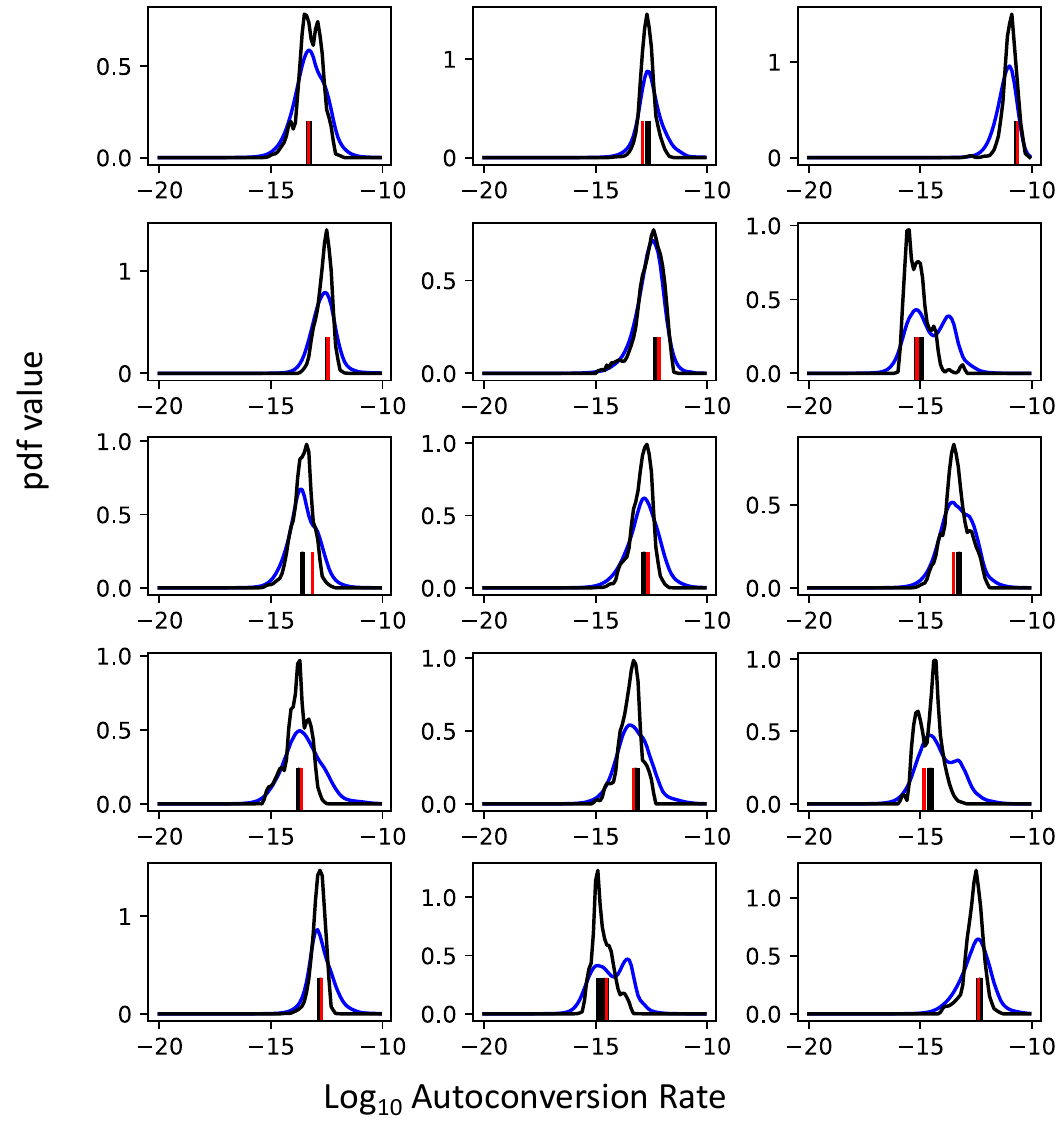}}
\caption{(continued)}
\label{fig: B2}
\end{figure}

 \begin{figure}[htp]
\centerline{\includegraphics[width=0.99\textwidth]{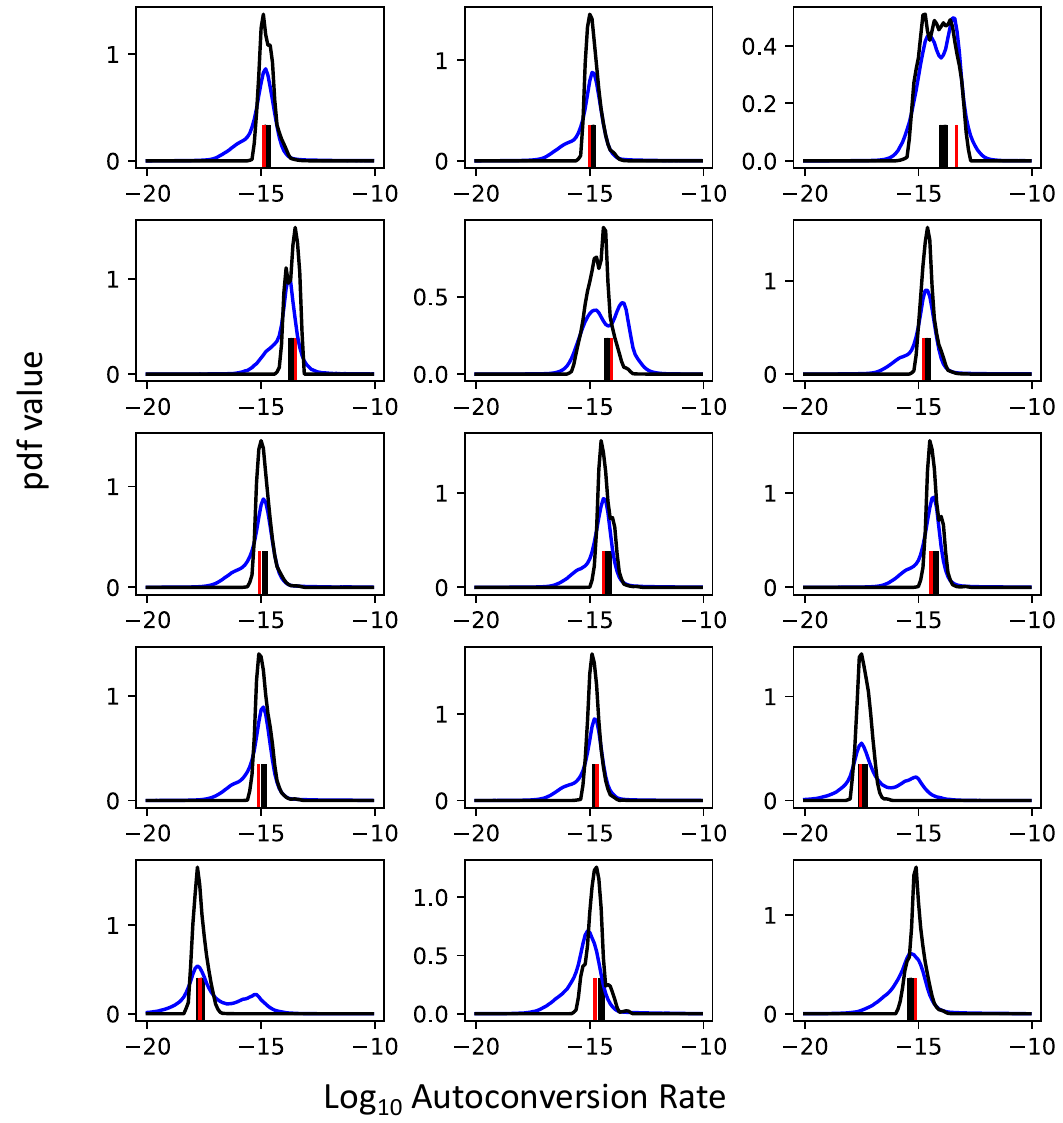}}
\caption{(continued)}
\label{fig: B3}
\end{figure}

 \begin{figure}[htp]
\centerline{\includegraphics[width=0.99\textwidth]{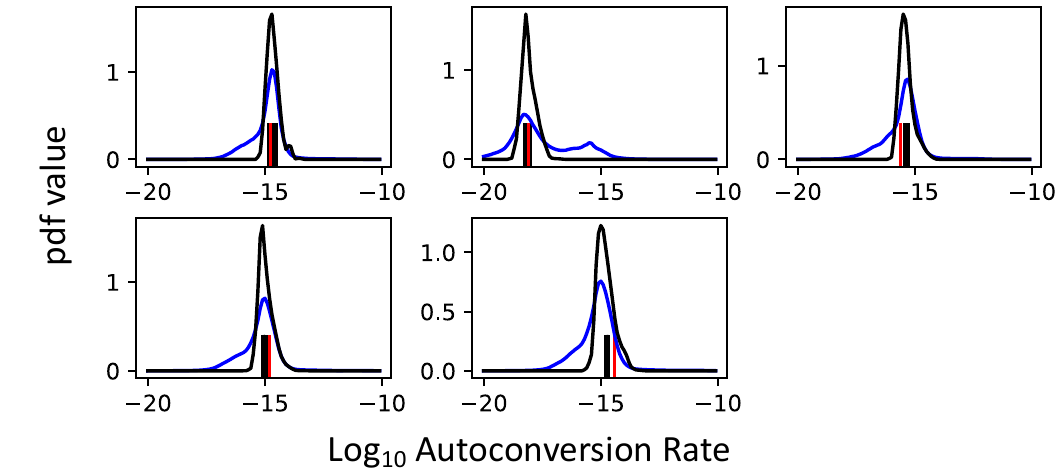}}
\caption{(continued)}
\label{fig: B4}
\end{figure}

\end{appendix}
\clearpage




\paragraph{Acknowledgments}
The authors thank the in-depth reviews by Dr. A. Carrassi and an anonymous reviewer, which have led to a much better manuscript.

\paragraph{Funding Statement}
Van Leeuwen was supported by the National Science Foundation under Grant 2220201. Chiu and Yang were supported by Department of Energy under Grant DE-SC0021167, and National Aeronautics and Space Administration Projects 80NSSC22K1546 and 80NSSC23K1017.

\paragraph{Data Availability Statement}
The training and testing data sets are available freely at \verb| https://doi.org/10.5281/zenodo.17623563.|

\paragraph{Author Contributions}
Conceptualization: PJvL and JCC, Methodology: PJvL, Data curation: JCC and CKY, Data visualisation: CKY and PJvL, Writing original draft: PJvL, All authors approved the final submitted draft.

\bibliographystyle{ametsocV6}
\bibliography{references}

\begin{thebibliography}{34}
\providecommand{\natexlab}[1]{#1}
\providecommand{\url}[1]{\texttt{#1}}
\renewcommand{\UrlFont}{\rmfamily}
\providecommand{\urlprefix}{URL }
\expandafter\ifx\csname urlstyle\endcsname\relax
  \providecommand{\doi}[1]{https://doi.org/\discretionary{}{}{}#1}\else
  \providecommand{\doi}{https://doi.org/\discretionary{}{}{}\begingroup
  \urlstyle{rm}\Url}\fi
\providecommand{\eprint}[2][]{\url{#2}}

\bibitem[{Abdar et~al.(2021)}]{Abdar2021}
Abdar, M., and Coauthors, 2021: A review of uncertainty quantification in deep
  learning: Techniques, applications and challenges. \textit{Information
  Fusion}, \textbf{76}, 243--297, \doi{10.1016/j.inffus.2021.05.008}.

\bibitem[{Ades and Van~Leeuwen(2012)Ades, and Van~Leeuwen}]{Ades2012}
Ades, M., and P.~J. Van~Leeuwen, 2012: An exploration of the equivalent-weights
  particle filter. \textit{Quarterly Journal of the Royal Meteorological
  Society}, \doi{10.1002/qj.1995}.

\bibitem[{Amini et~al.(2020)Amini, Schwarting, Soleimany,, and Rus}]{Amini2020}
Amini, A., W.~Schwarting, A.~Soleimany, and D.~Rus, 2020: Deep evidential
  regression. \doi{10.48550/arXiv.1910.02600}.

\bibitem[{Andrieu et~al.(2003)Andrieu, de~Freitas, Doucet,, and
  Jordan}]{Andrieu2003}
Andrieu, C., N.~de~Freitas, A.~Doucet, and M.~I. Jordan, 2003: An introduction
  to mcmc for machine learning. \textit{Machine Learning}, \textbf{50}, 5--43,
  \doi{10.1023/A:1020281327116}.

\bibitem[{Balasubramanian et~al.(2014)Balasubramanian, Ho,, and
  Vovk}]{Balasubramanian2014}
Balasubramanian, V., S.-S. Ho, and V.~Vovk, Eds., 2014: \textit{Conformal
  Prediction for Reliable Machine Learning, Theory, Adaptations and
  Applications}. Elsevier.

\bibitem[{Blundell et~al.(2015)Blundell, Cornebise, Kavukcuoglu,, and
  Wierstra}]{Blundell2015}
Blundell, C., J.~Cornebise, K.~Kavukcuoglu, and D.~Wierstra, 2015: Weight
  uncertainty in neural network. \textit{International Conference on Machine
  Learning}, 1613--1622.

\bibitem[{Breiman(1996)}]{Breiman1996}
Breiman, L., 1996: Bagging predictors. \textit{Machine Learning}, \textbf{24},
  \doi{10.1007/BF00058655}.

\bibitem[{Cheng et~al.(2023)Cheng, Quilodr{\'a}n-Casas, Ouala
  et~al.}]{Cheng2023}
Cheng, S., C.~Quilodr{\'a}n-Casas, S.~Ouala, and Coauthors, 2023: Machine
  learning with data assimilation and uncertainty quantification for dynamical
  systems: A review. \textit{IEEE/CAA Journal of Automatica Sinica},
  \textbf{10~(6)}, 1361--1387, \doi{10.1109/JAS.2023.123537}.

\bibitem[{Chiu et~al.(2021)Chiu, Yang, Van~Leeuwen, Feingold et~al.}]{Chiu2021}
Chiu, C.-C., C.-K. Yang, P.~J. Van~Leeuwen, G.~Feingold, and Coauthors, 2021:
  Observational constraints on warm cloud microphysical processes using machine
  learning and optimization techniques. \textit{Geophysical Research Letters},
  \doi{10.1029/2020GL091236}.

\bibitem[{Doucet et~al.(2001)Doucet, de~Freitas,, and Gordon}]{Doucet2001}
Doucet, A., N.~de~Freitas, and N.~Gordon, Eds., 2001: \textit{Sequential Monte
  Carlo Methods in Practice}. Statistics for Engineering and Information
  Science, Springer-Verlag New York, 581 pp.

\bibitem[{Evensen et~al.(2022)Evensen, Vossepoel,, and van
  Leeuwen}]{Evensen2022}
Evensen, G., F.~Vossepoel, and P.~J. van Leeuwen, 2022: \textit{Data
  Assimilation Fundamentals}. Springer, \doi{10.1007/978-3-030-96709-3}.

\bibitem[{Gal and Ghahramani(2016)Gal, and Ghahramani}]{Gal2016}
Gal, Y., and Z.~Ghahramani, 2016: Dropout as a bayesian approximation:
  Representing model uncertainty in deep learning. \textit{International
  Conference on Machine Learning}, Vol.~48, 1050--1059.

\bibitem[{Gawlikowski et~al.(2023)}]{Gawlikowski2023}
Gawlikowski, J., and Coauthors, 2023: A survey of uncertainty in deep neural
  networks. \textit{Artif. Intell. Rev.}, \textbf{56}, 1513--1589,
  \doi{10.1007/s10462-023-10562-9}.

\bibitem[{Gettelman et~al.(2021)Gettelman, Gagne, Chen, Christensen, Lebo,
  Morrison,, and Gantos}]{Gettelman2021}
Gettelman, A., D.~J. Gagne, C.-C. Chen, M.~W. Christensen, Z.~J. Lebo,
  H.~Morrison, and G.~Gantos, 2021: Machine learning the warm rain process.
  \textit{Journal of Advances in Modeling Earth Systems}, \textbf{13~(2)},
  e2020MS002\,268, \doi{https://doi.org/10.1029/2020MS002268}.

\bibitem[{Glienke and Mei(2019)Glienke, and Mei}]{Glienke2019}
Glienke, S., and F.~Mei, 2019: Two-dimensional stereo (2d-s) probe instrument
  handbook. Tech. rep., DOE ARM.
  \urlprefix\url{https://www.arm.gov/publications/tech_reports/handbooks/doe-sc-arm-tr-233.pdf}.

\bibitem[{Glienke and Mei(2020)Glienke, and Mei}]{Glienke2020}
Glienke, S., and F.~Mei, 2020: Fast cloud droplet probe (fcdp) instrument
  handbook. Tech. rep., DOE ARM.
  \urlprefix\url{https://www.arm.gov/publications/tech_reports/handbooks/doe-sc-arm-tr-238.pdf}.

\bibitem[{He et~al.(2025)He, Jiang, Xiao, Xu, ,, and Li}]{He2025}
He, W., Z.~Jiang, T.~Xiao, Z.~Xu, , and Y.~Li, 2025: A survey on uncertainty
  quantification methods for deep learning.
  Https://arxiv.org/pdf/2008.02627.pdf.

\bibitem[{Isaksen et~al.(2010)Isaksen, Bonavita, Buizza, Fisher, Haseler,
  Leutbecher,, and Raynaud}]{Isaksen2010}
Isaksen, L., M.~Bonavita, R.~Buizza, M.~Fisher, J.~Haseler, M.~Leutbecher, and
  L.~Raynaud, 2010: Ensemble of data assimilations at ecmwf. \textit{ECMWF
  Technical Memoranda}, Vol. 636, \doi{10.1007/978-3-319-46448-0_15}.

\bibitem[{Lakshminarayanan et~al.(2017)Lakshminarayanan, Pritzel,, and
  Blundell}]{Lakshminarayanan2017}
Lakshminarayanan, B., A.~Pritzel, and C.~Blundell, 2017: Simple and scalable
  predictive uncertainty estimation using deep ensembles. \textit{Advances in
  Neural Information Processing Systems}, Vol.~30.

\bibitem[{MacKay(1992)}]{MacKay1992}
MacKay, D.~J., 1992: A practical bayesian framework for backpropagation
  networks. \textit{Neural Computation}, \textbf{4}, 448--472,
  \doi{10.1162/neco.1992.4.3.448}.

\bibitem[{Mei et~al.(2020)Mei, Wang, Comstock et~al.}]{Mei2020}
Mei, F., J.~Wang, J.~M. Comstock, and Coauthors, 2020: Comparison of aircraft
  measurements during goamazon2014/5 and acridicon-chuva. \textit{Atmospheric
  Measurement Techniques}, \textbf{13}, 661--684,
  \doi{10.5194/amt-13-661-2020}.

\bibitem[{Neal(1996)}]{Neal1996}
Neal, R.~M., 1996: \textit{Bayesian Learning for Neural Networks}. Springer,
  \doi{10.1007/978-1-4612-0745-0}.

\bibitem[{Nix and Weigend(1994)Nix, and Weigend}]{Nix1994}
Nix, D.~A., and A.~S. Weigend, 1994: Estimating the mean and variance of the
  target probability distribution. \textit{IEEE International Conference on
  Neural Networks}, \doi{10.1109/ICNN.1994.374138}.

\bibitem[{Osband(2016)}]{Osband2016}
Osband, I., 2016: Risk versus uncertainty in deep learning: Bayes, bootstrap
  and the dangers of dropout. NIPS 2016 Workshop on Bayesian Deep Learning.

\bibitem[{Pfreundschuh et~al.(2018)Pfreundschuh, Eriksson, Duncan
  et~al.}]{Pfreundschuh2018}
Pfreundschuh, S., P.~Eriksson, D.~Duncan, and Coauthors, 2018: A neural network
  approach to estimating a posteriori distributions of bayesian retrieval
  problems. \textit{Atmospheric Measurement Techniques}, \textbf{11},
  4627--4643, \doi{10.5194/amt-11-4627-2018}.

\bibitem[{Sohl-Dickstein et~al.(2015)Sohl-Dickstein, Weiss, Maheswaranathan,,
  and Ganguli}]{Sohl2015}
Sohl-Dickstein, J., E.~Weiss, N.~Maheswaranathan, and S.~Ganguli, 2015: Deep
  unsupervised learning using nonequilibrium thermodynamics.
  \textit{Proceedings of Machine Learning Research}, Vol.~37, 2256--2265.

\bibitem[{S{\o}nderby et~al.(2020)S{\o}nderby, Espeholt, Heek
  et~al.}]{Sonderby2020}
S{\o}nderby, C.~K., L.~Espeholt, J.~Heek, and Coauthors, 2020: Metnet: A neural
  weather model for precipitation forecasting. Google research,
  arXiv:2003.12140v2.

\bibitem[{Van~Leeuwen(2010)}]{VanLeeuwen2010}
Van~Leeuwen, P.~J., 2010: Nonlinear data assimilation in geosciences: an
  extremely efficient particle filter. \textit{Quarterly Journal of the Royal
  Meteorological Society}, \textbf{136}, 1991--1996, \doi{10.1002/qj.699}.

\bibitem[{{Van Leeuwen}(2019)}]{VanLeeuwen2019}
{Van Leeuwen}, P.~J., 2019: Nonlocal observations and information transfer in
  data assimilation. \textit{Frontiers Appl. Math. Stats}, \textbf{26},
  \doi{10.3389/fams.2019.00048}.

\bibitem[{Van~Leeuwen(2020)}]{VanLeeuwen2020}
Van~Leeuwen, P.~J., 2020: A consistent interpretation of the stochastic version
  of the ensemble kalman filter. \textit{Quarterly Journal of the Royal
  Meteorological Society}, \doi{10.1002/qj.3819}.

\bibitem[{Van~Leeuwen et~al.(2015)Van~Leeuwen, Cheng,, and
  Reich}]{VanLeeuwen2015b}
Van~Leeuwen, P.~J., Y.~Cheng, and S.~Reich, 2015: \textit{Nonlinear Data
  Assimilation}. Springer, \doi{10.1007/978-3-319-18347-3}.

\bibitem[{Verdoja and Kyrki(2021)Verdoja, and Kyrki}]{Verdoja2021}
Verdoja, F., and V.~Kyrki, 2021: Notes on the behavior of mc dropout. Presented
  at ICML Workshop on Uncertainty and Robustness in Deep Learning,
  https://arxiv.org/pdf/2008.02627.pdf.

\bibitem[{Wang et~al.(2016)Wang, Li, Ouyang,, and Wang}]{Wang2016}
Wang, Z., H.~Li, W.~Ouyang, and X.~Wang, 2016: Learnable histogram: Statistical
  context features for deep neural networks. \textit{European Conference on
  Computer Vision}, 246--262, \doi{10.1007/978-3-319-46448-0_15}.

\bibitem[{Yu and Ma(2021)Yu, and Ma}]{Yu2021}
Yu, S., and J.~Ma, 2021: Deep learning for geophysics: Current and future
  trends. \textit{Reviews of Geophysics}, \textbf{59},
  \doi{10.1029/2021RG000742}.

\end{thebibliography}


\end{document}